\pdfoutput=1

\documentclass[11pt]{article}

\newif\ifcomment\commenttrue{}
\newif\ifcomment\commentfalse{}

\usepackage[a-1b]{pdfx}

\usepackage{framed}
\usepackage{mdwlist}
\usepackage{siunitx}
\usepackage{latexsym}
\usepackage{colortbl}
\usepackage{xcolor}
\usepackage{nicefrac}
\usepackage{booktabs}
\usepackage{fnpct}
\usepackage{amsfonts}
\usepackage[T1]{fontenc}
\usepackage{bold-extra}
\usepackage{amsmath}
\usepackage{amssymb}
\usepackage{bm}
\usepackage{graphicx}
\usepackage{mathtools}
\usepackage{microtype}
\usepackage{multirow}
\usepackage{multicol}
\usepackage{placeins}
\usepackage{xpatch}
\usepackage{latexsym,comment}
\usepackage[normalem]{ulem}
\hypersetup{
    colorlinks,
    linkcolor={red!!black},
    citecolor={blue!50!black},
    urlcolor={blue!80!black}
}

\newcommand*{\missingreference}{{\colorbox{red}{?reference?}}}
\newcommand*{\missingcitation}{{\colorbox{red}{?citation?}}}

\makeatletter
\xpatchcmd{\@setref}{\bfseries}{\missingreference}{}{}
\def\@citex[#1]#2{\leavevmode
    \let\@citea\@empty
    \@cite{\@for\@citeb:=#2\do
        {\@citea\def\@citea{,\penalty\@m\ }%
            \edef\@citeb{\expandafter\@firstofone\@citeb\@empty}%
            \if@filesw\immediate\write\@auxout{\string\citation{\@citeb}}\fi
            \@ifundefined{b@\@citeb}{\hbox{\reset@font\missingcitation}%
                \G@refundefinedtrue
                \@latex@warning
                {Citation `\@citeb' on page \thepage \space undefined}}%
            {\@cite@ofmt{\csname b@\@citeb\endcsname}}}}{#1}}
\makeatother

\newcommand{\feat}[1]{{\small \texttt{#1}}}

\newcommand{\gem}[1]{\mbox{\textsc{gem}}}
\newcommand{\abr}[1]{\textsc{#1}}

\newcommand{\g}{\, | \,}

\newcommand{\textvtt}[1]{{\normalfont\fontfamily{cmvtt}\selectfont #1}}

\renewenvironment{quote}
{\list{}{\rightmargin\leftmargin}%
    \item\relax\small\ignorespaces}
{\unskip\unskip\endlist}

\DeclareMathOperator*{\argmin}{arg\,min}

\newcommand{\hidetext}[1]{}
\newcommand{\ignore}[1]{}

\ifcomment
    \newcommand{\pinaforecomment}[3]{\colorbox{#1}{\parbox{.8\linewidth}{#2: #3}}}

    \newcommand{\prtodo}[1]{\pinaforecomment{lightblue}{pr}{#1}}
    \newcommand{\prtodoi}[1]{\pinaforecomment{lightblue}{pr}{#1}}
    \newcommand{\baseinlinecomment}[1]{{\textcolor[rgb]{1.0, 0.0, 0.0}{#1}}}
\else
    \newcommand{\baseinlinecomment}[1]{}
    \newcommand{\pinaforecomment}[3]{}
    \newcommand{\prtodo}[1]{}
    \newcommand{\prtodoi}[1]{}
\fi

\definecolor{lightblue}{HTML}{3cc7ea}
\definecolor{CUgold}{HTML}{CFB87C}
\definecolor{grey}{rgb}{0.95,0.95,0.95}
\definecolor{ceil}{rgb}{0.57, 0.63, 0.81}
\definecolor{UMDred}{HTML}{ed1c24}
\definecolor{UMDyellow}{HTML}{ffc20e}
\definecolor{darkgreen}{rgb}{0.0, 0.5, 0.0}
\definecolor{greencustom}{rgb}{0.0, 0.4, 0.0}
\definecolor{lightred}{rgb}{1.0, 0.4, 0.4}
\definecolor{lightgreen}{rgb}{0.6, 1.0, 0.6}
\definecolor{britishracinggreen}{rgb}{0.0, 0.26, 0.15}

\newcommand{\jbgcomment}[1]{\pinaforecomment{lightred}{JBG}{#1}}
\newcommand{\mgorcomment}[1]{\pinaforecomment{lightgreen}{mgor}{#1}}

\newcommand{\rightcomment}[2]{\baseinlinecomment{$\bm{\left[\right.}$\textbf{#1}: {#2}$\bm{\left.\right]\rightarrow$}}}

\newcommand{\mgor}[1]{\mgorcomment{#1}}
\newcommand{\mgr}[1]{\rightcomment{mgor}{#1}}

\newcommand{\smallurl}[1]{ \begin{tiny}\url{#1}\end{tiny}}

\newcommand{\ai}[0]{\abr{ai}}
\newcommand{\caimirafull}[0]{Content-aware, Identifiable, and Multidimensional Item Response Analysis}
\newcommand{\caimira}[0]{\abr{caimira}}
\newcommand{\mirt}[0]{\abr{mirt}}
\newcommand{\irt}[0]{\abr{irt}}
\newcommand{\llm}[1]{\abr{llm}#1}
\newcommand{\qb}[0]{\abr{qb}}
\newcommand{\pb}[0]{ProtoBowl}
\newcommand{\qa}[0]{\abr{qa}}

\newcommand{\nlp}[0]{\abr{nlp}}

\newcommand{\bert}{\abr{bert}}

\usepackage{style/acl}
\usepackage{lmodern}
\usepackage{times}
\usepackage{bookmark}
\usepackage[T1]{fontenc}

\usepackage[utf8]{inputenc}

\usepackage{microtype}
\usepackage[most]{tcolorbox} 
\usepackage{tabularx}
\usepackage{lipsum} 

\newcommand{\datasetsize}[0]{$3042$}

\newcommand{\gptthree}[0]{\abr{gpt-{\small3}}} 
\newcommand{\gptfour}[0]{\abr{gpt-{\small4}-turbo}} 
\newcommand{\gptfouro}[0]{\abr{gpt-{\small4o}}} 
\newcommand{\llama}[0]{\abr{llama-{\small3}-{\small70}b}} 
\newcommand{\bmtfive}[0]{\abr{bm{\small25}}} 

\newcommand{\disc}[1]{\ensuremath{\bm{\alpha_{#1}}}}        
\newcommand{\skill}[1]{\ensuremath{\mathbf{s_{#1}}}}        
\newcommand{\diff}[1]{\ensuremath{\mathbf{d_{#1}}}}         
\newcommand{\ediff}[1]{\ensuremath{\mathbf{d^{(e)}_{#1}}}}  
\newcommand{\rel}[1]{\ensuremath{\mathbf{r_{#1}}}}          
\newcommand{\qbid}[1]{{\small\texttt{#1}}}                  

\definecolor{qbred}{rgb}{0.5, 0.0, 0.0}
\definecolor{agentblue}{HTML}{3366CC}
\definecolor{agentgreen}{HTML}{109618}
\definecolor{agentpurple}{HTML}{990099}
\definecolor{agentlightblue}{HTML}{0099C6}
\definecolor{agentpink}{HTML}{DD4477}
\definecolor{agentdarkblue}{HTML}{316395}
\definecolor{caimirayellow}{HTML}{E69F00}
\definecolor{caimiradarkyellow}{HTML}{CEA146}
\definecolor{caimiragreen}{HTML}{509364}
\definecolor{caimiradarkgreen}{HTML}{3c6c4c}
\definecolor{dullred}{HTML}{953131}
\definecolor{dullblue}{HTML}{175B98}

\title{Do great minds think alike? Investigating Human-AI Complementarity \\ in Question Answering with \caimira{}~\includegraphics[height=0.04\textwidth]{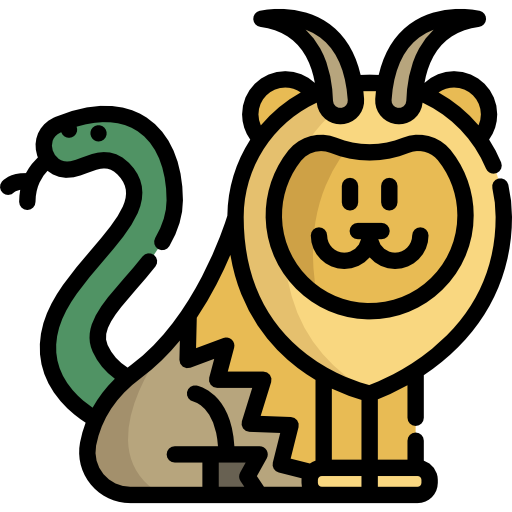} }

\author{%
  Maharshi Gor$^1$ \qquad Hal Daumé III$^{1, 2}$ \qquad Tianyi Zhou$^1$ \qquad Jordan Boyd-Graber$^1$ \\ \\
  $^1$University of Maryland \qquad $^2$Microsoft Research \\
  \texttt{mgor@cs.umd.edu}
}

\begin{document}

\maketitle

\begin{abstract}

Recent advancements of large language models (\llm{s})
have led to claims of \abr{ai} surpassing humans
in natural language processing (\nlp) tasks such as textual understanding and reasoning.
This work investigates these assertions by introducing \caimira{}, a novel framework
rooted in item response theory (\abr{irt}) that enables quantitative assessment and
comparison of problem-solving abilities of question-answering (\qa{}) agents: humans
and \ai{} systems.
Through analysis of over 300,000 responses from \textasciitilde~70 \ai{} systems
and 155 humans across thousands of quiz questions, \caimira{} uncovers distinct
proficiency patterns in knowledge domains and reasoning skills. 
Humans outperform \ai{} systems in knowledge-grounded abductive and conceptual reasoning,
while state-of-the-art \llm{s} like \gptfour{} and \llama{} show superior performance on
targeted information retrieval and fact-based reasoning, particularly when information gaps
are well-defined and addressable through pattern matching or data retrieval.
These findings highlight the need for future \qa{} tasks to focus on questions that
challenge not only higher-order reasoning and scientific thinking, but also demand
nuanced linguistic interpretation and cross-contextual knowledge application,
helping advance \ai{} developments that better emulate or complement human cognitive
abilities in real-world problem-solving.
\end{abstract}

\section{Introduction}\label{sec:intro}
The \abr{nlp} community has focused on
human behavior \textit{emulation}, treating human performance as ceiling for models.
However, the latest wave of \llm{s} has turned the discussion to supremacy: models are
purportedly acing tests~\cite{liu2023evaluating, hendrycks2020measuring} that many humans
find challenging.\footnote{As should hopefully be clear from the rest of the paper,
we are highly dubious of these claims, particularly on multi-choice tests with
copious study material online. But this is outside the main scope of \emph{this} paper.}
%

A notable 2010 example was \abr{ibm} Watson's \textit{tour de force}
performance~\citet{ferruci-10} on \textit{Jeopardy!}.
While Watson defeated the two humans on stage over a few dozen questions,
a thorough, quantitative examination of the relative strengths
and weaknesses of human vs.\ computer on question answering (\qa{}),
particularly with the new panoply of recent \llm{s}, remains absent.
%

To address this gap, we turn to Item Response Theory (\irt{}, \S\ref{subsec:irt-review}),
a statistical framework, originally developed in psychometrics~\citep{Santor1998ProgressIT},
used for constructing effective standardized tests, by modeling the interaction between
individuals and test items (questions).
\irt{} is particularly suited for our analysis because it allows
us to simultaneously assess the abilities of respondents (in our
case, both humans and \ai{} systems) and the characteristics of
test items (our questions).
This dual assessment is crucial for understanding the nuanced
differences in performance between humans and \ai{} systems across various types of questions.
\begin{figure}[t]
    \centering
    \includegraphics[width=1.0\linewidth]{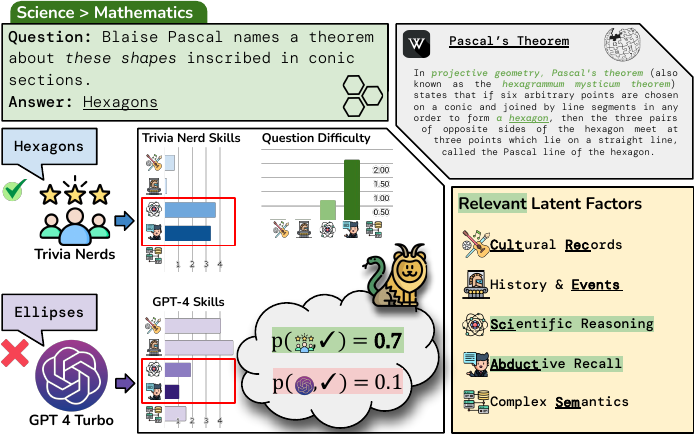}
    \caption{Response Correctness prediction using Agent skills and Question difficulty 
    over relevant latent factors. We list the five latent factors that
    \caimira{} discovers, and highlight the relevant ones (green), which
    contribute to estimating whether an agent will respond to the example
    question correctly. The agent skills over these relevant factors are highlighted in red boxes.}\label{fig:teaser}
  \end{figure}

\jbgcomment{Suggestion:
Existing IRT models use hand-crafted features (cite LLTM) or use
representations that are independent of correctness prediction.
}

Building upon \irt{}, we introduce \caimira{}---\caimirafull{}
(pronounced \textbf{\textit{Chimera}}~\includegraphics[height=0.025\textwidth]{figures/chimera.png})---a neural framework\footnote{The implementation can
be found at \url{https://github.com/maharshi95/neural-irt}} that overcomes key challenges
of applying \irt{} to \qa{}. \caimira{} uses question text to infer characteristics,
enabling generalization to new questions without needing prior responses.


For our questions, we use a \qa{} format~\citep[QuizBowl]{boyd-graber-12}
specifically designed for effective comparison between \qa{} agents (\S~\ref{subsec:qb}).
We then apply \caimira{}~(\S~\ref{sec:qanda}) to responses collected from
155 human trivia players, and a wide range (\textasciitilde~70) of \qa{} systems,
over thousands of these carefully crafted questions
that probe knowledge recall and reasoning capabilities.
\caimira{} uncovers latent aspects~(\autoref{fig:relevance-analysis})
that encapsulate different knowledge domains and reasoning skills, that best contrast
agents' capabilities.

Humans and \qa{} systems' skills are strikingly different across these latent axes
(\autoref{fig:agent-skills}).
Human responses reflect their superior interpretative abilities, instinctive thinking,
and cognitive flexibility.
\textbf{This is particularly evident in questions demanding conceptual and knowledge-grounded
abductive reasoning, characterized by indirect narrative references and ambiguous
information gaps, where humans make intuitive leaps and draw connections that may
not be immediately apparent.}
Conversely, large-scale \llm{s} like \gptfour{} and \llama{} demonstrate superior ability
in retrieving specific information about events and locations, outdoing
humans on questions loaded with entity-specific details---a feat we attribute
to their extensive parametric memory.
\caimira{} also reveals questions that, while easily matched to relevant documents
by retrieval systems, challenge most \llm{s} in extracting the
final answer. 
These questions feature complex sentence structures and semantic relationships,
that turn simple information retrieval into demanding reading comprehension.

In conclusion, this study provides insights into the strengths and weaknesses of
human and \ai{} question answering, laying the groundwork for future \ai{}
developments that better emulate or complement human cognitive abilities.
In doing so, it underscores the need for sophisticated benchmarks to controllably
distinguish between proficient and less capable \qa{} systems, especially in
areas demanding deeper, conceptual, and linguistic understanding.

\mgor{
A framework like \caimira{} can be used to i\) to diagnose the (lacking) skills of \qa{} agents via their incorrect responses, ii\) discover a smaller and yet equally effective set of questions to assess the differentiating capabilities of \qa{} agents, in case the full set is too large or too expensive to run in a real-world scenario, and iii\) Design a routing system for directing questions to the most appropriate \qa{} agents based on their skills for effective high-quality annotation collection.
}

\section{Background and Preliminaries}\label{sec:background}
This section describes the source of the Quizbowl \qa{} data~(\S~\ref{subsec:qb}) and preliminaries of \irt{} and \mirt{}~(\S~\ref{subsec:irt-review}),
the foundation of \caimira{}~(\S~\ref{sec:kira}).
%

\subsection{\abr{Quizbowl}: Where Trivia Nerds Practice}\label{subsec:qb}
Our overarching goal is to identify similarities and differences between 
how systems and humans respond to questions.
These questions must be \textit{diverse}, less prone to false presuppositions,
and designed to be challenging for humans, enabling us to draw conclusions
about the strengths and weaknesses of agents without needing to
``question the question''~\cite{min2020ambigqa, yu2022crepe}.
Following the categorization by \citet{rogers-21}, we focus on depth-testing
``probing'' questions over ``information seeking'' ones. This approach aligns with the 
Manchester paradigm outlined by \citet{rodriguez-boyd-graber-2021-evaluation}, which highlights 
the significance of research agendas in the development of human-like, intelligent \qa{} systems.
\begin{figure}[t]
    \centering
    \includegraphics[width=1.0\linewidth]{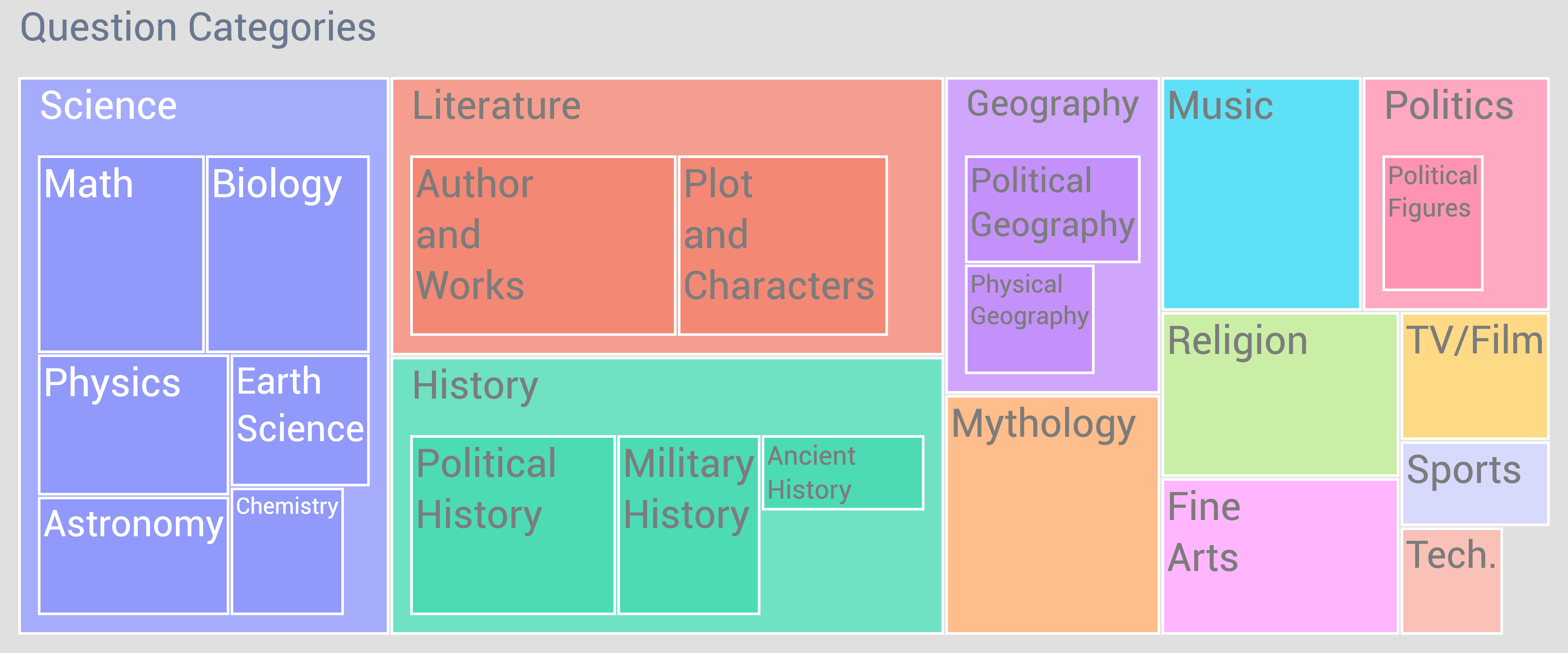}
    \caption{Distribution of question categories and subcategories over our
    dataset of 3042 questions.}\label{fig:question-categories}
  \end{figure}
More importantly, we need questions with many examples of diverse human answers.
While humans may not answer Google queries~\citep{kwiatkowski-19} for
fun, they do answer trivia questions as a hobby or to prepare for trivia competitions.
Hence, we use the ``Protobowl''~\cite{he-16}, a dataset of trivia questions based
on the Quizbowl (\qb{}) \abr{qa} setting~\cite{boyd-graber-12}.
Quizbowl, the source of questions for \pb{}, is a trivia game consisting of questions
with sentence-clues decreasing in difficulty and culminating with a ``giveaway'' hint at the
end of the question. 
It is the only open source \qa{} dataset that
contains records of many human players of varying levels of expertise
answering questions across different categories like history, science
and literature\footnote{\autoref{appendix:qb} provides further details into the
\qb{} dataset.} (\autoref{fig:question-categories}).
%

\subsection{A review of Item Response Theory (\irt)}\label{subsec:irt-review}

We compare humans and \ai{} systems by capturing their
skills using Item Response Theory (\irt{}), 
a framework used to understand question quality and participant strengths,
by analyzing responses (ruled as correct or incorrect)
to a set of questions (or, ``items''). 
It is widely adopted in psychometrics~\citep{morizot2009toward},
medical education~\citep{downing2003item}, 
and other fields for developing standardized tests for human subjects.

In the context of \emph{this} work, \irt{} assumes
(1) a set of question-answer pairs,
(2) subjects spanning humans and \qa{} systems, and
(3) binary correctness rulings of their responses.
The \irt{} objective is to predict the response correctness ($U_{i,j}$)
based on the subject's skill~$s_i$ and
the question's difficulty~$d_j$, where $i$ and $j$ are the indices
of the subject and question, respectively.
The probability of response correctness, $p(U_{i, j} = 1)$, is modeled as $\sigma(s_i - d_j)$, where $\sigma$ is the sigmoid function.
\begin{equation}\label{equ:irt}
    p(U_{i, j} = 1 \g s_i, d_j) = \sigma(s_i - d_j).
\end{equation}
The learning objective is to model skill and difficulty parameters that
best fit assumed priors, given observed response data, typically using Bayesian inference.
Existing \irt{} applications in \nlp{} often model item characteristics in
one dimension~\citep{lalor2019learning}, assuming a linear hierarchy in difficulty
and skill levels. This approach is limiting when distinguishing between agents
in \nlp{} tasks. For example, if a history question $q_h$ is found to be
more difficult than a science question $q_s$ ($d_h > d_s$), the model asserts that agents
correctly answering $q_h$ also correctly answer $q_s$, and vice versa.
\paragraph{Multidimensional Latent IRT (\mirt{}).}
To relax the monotonicity assumption and model multi-factor
characteristics, \mirt{} was developed~\citep{reckase2006mirt, chalmers2012mirt}.
It models two question characteristics: a scalar \textit{difficulty} $d_j$, and an
$m$-dimensional discriminability $\disc{j}$ that interacts with the $m$-dimensional
\textit{skill} vector $\skill{i}$. The skill value $\skill{i,k}$ corresponds to the
agent's expertise on the $k^\text{th}$ \emph{latent aspect}.
The objective then becomes:
\begin{align}\label{equ:mirt}
    p(U_{i, j} = 1 \g \skill{i}, d_j, \disc{j}) = \sigma(\skill{i}^\intercal\disc{j} -  d_j).
\end{align}
The discriminability $\disc{j}$ captures how sensitively the correctness
probability changes with each dimension of the agent skill $\skill{i}$.
%
To mitigate overexpressibility, \mirt{} assumes $\disc{j}$ to have a gamma prior,
allowing only positive values. But, non-identifiability issues~\citep{10.1093/bioinformatics/btp358}
persist.\footnote{Negative skill values ($\skill{i} < 0$) and their interaction with $\disc{j}>1$
could mimic similar likelihood estimates ($p(U_{i, j})$) as that of positive skills
($\skill{i} > 0$) with $\disc{j} > 1$.}
A common practice of using hierarchical priors for resolving this makes optimization
unstable for higher dimensions.
%
Lastly, the model's exclusive dependence on question identifiers (\qbid{q31\_2})
treats questions as unrelated and hinders generalization.
The characteristics learned this way do not identify the difference in the questions
based on their content~\cite{rodriguez-etal-2022-clustering}
%

\section{Bootstrapping \irt{} with \caimira{}}\label{sec:kira}
\begin{figure*}
        \centering
        \includegraphics[width=\linewidth]{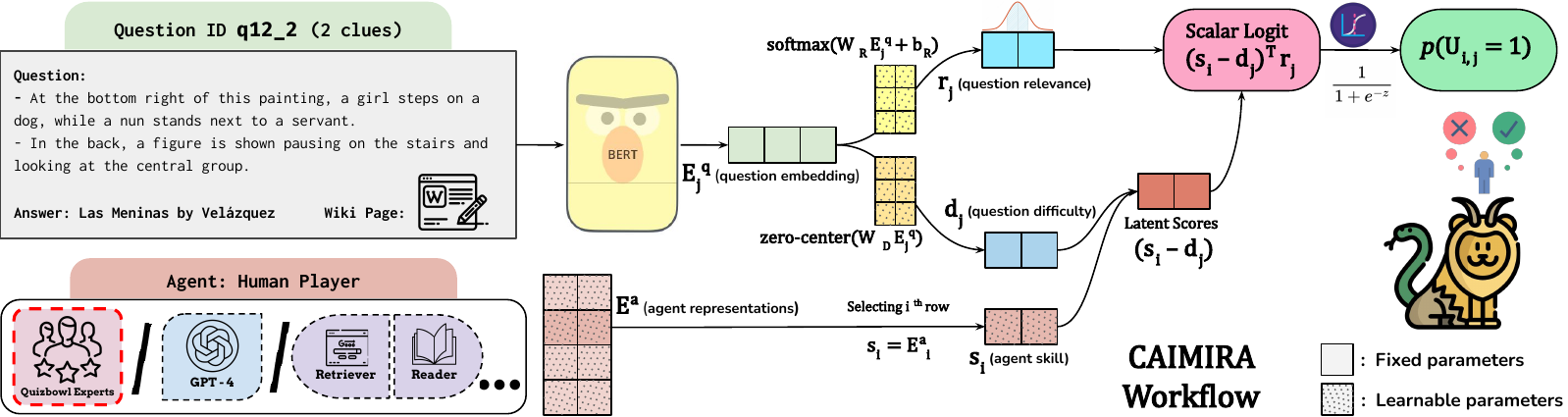}
        \caption{The \caimira{} workflow. It predicts the probability of agent-$i$ correctly 
        answering question-$j$ using a model in Eq.~\eqref{equ:kira}. Here, the question's 
        raw relevance $\mathbf{r'_j}$ and raw difficulty $\mathbf{d;_j}$ are multidimensional
        and computed by learnt linear transformations over the question embedding $\mathbf{E}^q_j$ 
        (\S~\ref{subsec:equations}), and the agent skill $\mathbf{s_i}$ is extracted from 
        a learnable agent embedding matrix $\mathbf{E}^a$. $\mathbf{r_j}$ is a
        probability distribution computed from the  raw reference $\mathbf{r'_j}$ and
        improves the interpretability of the multidimensional model (\S~\ref{sec:relevance});
        $\mathbf{d_j}$ is achieved by zero centering of the raw difficulty $\mathbf{d'_j}$,
        which addresses the non-identifiability issue of $\mathbf{s_i}$ and $\mathbf{d_j}$ in
        $(\mathbf{s_i}-\mathbf{d_j})$ (\S~\ref{subsec:zero-mean}).}\label{fig:workflow}
    \end{figure*}

We propose \caimira---\caimirafull{}, an \irt{} framework that addresses the limitations of \mirt{}
(\S~\ref{subsec:irt-review}) by introducing three key modifications:
(i) a novel concept of relevance (\rel{j}) for each item $j$,
(ii) zero-centered difficulty (\diff{j}), and
(iii) learnable content-aware transformations ($f_R$ and $f_D$)
that produce $\rel{j}$ and $\diff{j}$ from the raw questions.
These enable \caimira{} to provide interpretable and identifiable results,
and handle new questions without prior response data.
The response prediction model, the probability of agent $i$ correctly answering question $j$,
for an $m$-dimensional \caimira, is given by \autoref{equ:kira}.
%
\begin{align}\label{equ:kira}
    &p(U_{i, j} = 1 \g \skill{i}, \rel{j}, \diff{j}) = \sigma\left({(\skill{i} -  \diff{j})}^{\intercal} \rel{j}\right). \\
    &\text{\footnotesize where, $\skill{i} \in \mathbb R^m$ is agent skills, } \nonumber \\
    &\text{\footnotesize and, $\rel{j}, \diff{j}\in\mathbb R^m$ are question relevance and difficulty resp.} \nonumber
\end{align}
%

\subsection{Introducing question \emph{relevance} $\rel{j}$}\label{sec:relevance}
An \textit{interpretable} item response analysis should include
an item characteristic for each question that has the single responsibility of capturing
how relevant each latent aspect is for estimating the likelihood of
an agent correctly answering a particular question, $p(U_{i, j})$.
We call this \emph{relevance}.

Relevance \rel{j} measures how differences between and agent skills and
question difficulty ($\skill{i} - \diff{j}$), or \emph{latent scores},
align across the $m$-dimensions (Eq~\ref{equ:kira}),
assigning each dimension (or, latent aspect) a proportion (\rel{j,k}) to show its importance.
To ensure clarity and prevent overlap with \emph{difficulty}, \rel{j} is defined as a
probability distribution across the $m$ dimensions.
For instance, for a Thermodynamics question,
\caimira{} assigns greater relevance to dimensions
capturing physics knowledge and analytical reasoning, down weighing unrelated
dimensions like history or language.
This targeted aggregation of differences across relevant dimensions ensures that the
likelihood estimate $p(U_{i, j} = 1 \g \skill{i}, \rel{j}, \diff{j})$,
is both precise and contextually appropriate.

\paragraph{Connection to Topic Models}
This admixture mirrors the per-document allocation in topic models;
in \caimira{}, questions are admixtures of latent
aspects, or dimensions, with \emph{relevance} \rel{j} indicating
each dimension's contribution to the question. 
%
\vspace{-\baselineskip}
\subsection{Zero Centering of \emph{difficulty} $\diff{j}$}\label{subsec:zero-mean}
Aggregating \emph{differences} between agent skills and question difficulty
$(\skill{i} -  \diff{j})$ across dimensions (Eq~\ref{equ:kira}),
leads to \textit{non-unique} skill and difficulty values
for same likelihood estimate $p(U_{i, j} = 1)$.
We alleviate this non-identifiability issue by normalizing each question's
\textbf{raw difficulty} $\mathbf{d'_j}$ to have a zero mean for each
dimension~(Equation~\ref{eq:final-char}).
This normalization constrains skill and difficulty ranges
and enables comparisons across dimensions.

\subsection{Content-Aware Transformations}\label{subsec:equations}
\mgr{Motivate this by giving examples, either here or in Section~\ref{subsec:irt-review}}
\caimira{} improves upon \mirt{} by incorporating question content, enabling \caimira{}
to compute characteristics for new questions without requiring prior response data,
making it ``cold-start friendly''.
At its core, \caimira{} maps question text into
\emph{relevance} and \emph{difficulty} values using learnable
functions, $f_R, f_D: Q \to \mathbb{R}^m$,
transforming a question $q_j$ from the space of question texts $Q$
into raw relevance ($\mathbf{r'_j}$) and raw difficulty ($\mathbf{d'_j}$) vectors
(Figure~\ref{fig:workflow}).
These are modeled as linear transformations over a pre-trained embedder
$f_E: Q \to \mathbb{R}^n$ (e.g., \bert{}), which represents $q_j \in Q$
in an $n$-dimensional space as an embedding $\mathbf{e_j}$:
\vspace{-2mm}
\begin{align}
\mathbf{e_j} &:= f_E(q_j) = \texttt{BERT}(q_j), \\
\mathbf{r'_j} &:= f_R(q_j) = \mathbf{W}_R\ \mathbf{e_j} + \mathbf{b}_R, \\
\mathbf{d'_j} &:= f_D(q_j) = \mathbf{W}_D\ \mathbf{e_j} \label{eq:raw-char}
\end{align}
where $\mathbf{W}_R, \mathbf{W}_D \in \mathbb{R}^{m \times n}$
and $\mathbf{b}_R \in \mathbb{R}^{m}$ are the parameters of the linear transformations.
\footnote{We skip the bias term for $\mathbf{d'_j}$ since it is mean-centered.}
The raw values are then normalized to obtain final relevance ($\mathbf{r_j}$) and
difficulty ($\mathbf{d_j}$) values:
\vspace{-2mm}
\begin{align}
\rel{j} := \text{softmax}(\mathbf{r'_j}), \ \ \ & \diff{j} := \mathbf{d'_j} - \frac{1}{n_q}\sum_{j=1}^{n_q} \mathbf{d'_j}, \label{eq:final-char}
\end{align}
where $n_q$ is the number of questions in the dataset.
$\text{softmax}$ normalization for relevance ensures that the values sum to 1 across
$m$-dimensions, reflecting the relative importance of each latent aspect.
%

\paragraph{Agent Skills.}
\caimira{} learns an agent skill embedding matrix
$\mathbf{E}^a \in \mathbb{R}^{n_a \times m}$, where $n_a$ is the number of agents,
and the skill vector for agent $i$ is the $i^\text{th}$ row of this matrix:
\vspace{-2mm}
\begin{equation}
\vspace{-2mm}
\skill{i} = \mathbf{E}^a_i \label{eq:agent-skill}
\end{equation}
This approach allows \caimira{} to learn a compact representation of each agent's skills and question characteristics (difficulty and relevance),
across $m$ dimensions, which can be directly used in the response prediction model~(\autoref{equ:kira}).

\paragraph*{Learning Objective.}
To optimize \caimira{}'s parameters ($\Theta$), which include the agent skill embedding
matrix $\mathbf{E}^a$ and the linear transformation parameters
$\mathbf{b}_R$, $\mathbf{W}_R$ and $\mathbf{W}_D$, we use \emph{maximum a posteriori} estimate
(\abr{MAP}) based loss, which imposes implicit priors on the question characteristics and agent skills.
This combines a cross-entropy loss $\mathcal{L}_{\text{CE}}$~(Eq~\ref{eq:loss-ce}) with
regularization terms (Eq~\ref{eq:loss-re}):
\begin{align}
    \mathcal{L}_{\abr{ce}} &= -\dfrac{1}{N}\sum_{i,j} \ell_{\abr{ce}}(U_{i, j}, p(U_{i, j} = 1)), \label{eq:loss-ce} \\
    \mathcal{L}_{\text{reg}} &= \lambda_d \sum_{j}\| \diff{j}\|_1  + \lambda_s \sum_{i}\|\skill{i}\|_1, \label{eq:loss-re}
\end{align}
where $\ell_{CE}(x, y)$ is the cross-entropy loss between the true label $x$ and the
predicted probability in Eq.~\eqref{equ:kira}, $y$. $\|\cdot\|_1$ denotes the $\ell_1$ norm, and $\lambda_d$ and
$\lambda_s$ are the regularization hyperparameters.
Finally,
\begin{align}
    \mathcal{L}_{\caimira} &= \mathcal{L}_{\abr{ce}} + \mathcal{L}_{\text{reg}}, \label{eq:loss-caimira} \\
    \Theta_{\caimira} &\triangleq \argmin_{\Theta} \mathcal{L}_{\caimira} \label{eq:opt-caimira}
\end{align}


\section{Experimental Setup}\label{sec:qbqa}
This section describes how we collect responses from humans and \qa{} systems,
assess their answers, and analyze the latent traits learned by \caimira{}.


\paragraph*{Protobowl Logs.}
We collect player logs from the ``Protobowl'' platform over \qb{} questions spanning various
categories.~(Figure~\ref{fig:question-categories})
Player logs record question metadata, including category (e.g. History),
time taken to answer the question, answer string,
and the correctness ruling by the platform.
The best players have deep knowledge and excellent lateral thinking
skills~\citep{jennings-06}.

\paragraph*{Constructing QA Dataset.}
\qb{} questions are inherently multi-sentence (typically five) with each sentence serving as a
distinct clue for the answer.
In our dataset, each item is formed by cumulatively adding clues from a \qb{} question,
with the first item containing the initial clue and subsequent items
incorporating an additional clue each; i.e.,
the first item consists of only the first clue, the second item
comprises the first two clues together, and so on. 
This cumulative clue addition provides insight into how progressively
revealing information affects agents' response accuracy.

\paragraph{Mapping Player Responses to Cumulative Clues.}
Player responses are mapped to these cumulative clue items to analyze the
effectiveness of each clue set in eliciting correct answers.
Responses to {\qbid{q31}} after only the first clue are recorded
under {\qbid{q31\_1}}, and responses after the second clue
(which include the information from both clues) are recorded
under {\qbid{q31\_2}}, and so on.
This mapping is further refined through a backfilling process.
Because clues are meant to be progressively easier, we assume that a player who
correctly answers a question at clue $t$, would also correctly answer the question at clue $t^\prime > t$.
So, we mark those as correct as well. An analogous argument holds for $t^\prime < t$ when humans answer incorrectly.
Consequently, we collect a total of \datasetsize~entries in our refined dataset.\footnote{The dataset
is available on the HuggingFace platform as \href{https://huggingface.co/datasets/mgor/protobowl-11-13}{\texttt{mgor/protobowl-11-13}}.}

%


\mgorcomment{Comment: J!Archive can be useful for extracting more human data.}

\subsection{Human Agents}\label{subsec:human-agents}
In exploring the complementary \qa{} abilities of human and \ai{},
a key challenge is the sparsity of individual human data:
most players only engage with a set of few dozen questions. 
To address this, we form synthetic human agents by grouping
individual human players. 
This approach serves two primary purposes: it helps in accumulating a dataset
where agents have attempted a substantial portion of the questions, and it
mitigates the issue of non-representativeness of data from a few power users.

\paragraph*{Group Formation and Decision Mechanism}
Our dataset comprises only five human players who have answered over 1500 questions each.
While these ``power users'' are invaluable, relying solely on their data could
skew the understanding of human-AI interaction, as they might not be
representative of the broader player base.
Therefore, we introduce ``grouped human agents''. 
Each grouped agent is a synthetic construct, amalgamating responses
from multiple human players with similar skill levels.
We group human players such that the overall coverage of questions attempted by the group is maximized.
In cases where multiple players in a group answer the same question, we use a majority rule to determine the group's response.
If no majority is reached, a response is sampled based on the votes.\footnote{ 
This method is a basic approach to represent group decision-making,
acknowledging more complex dynamics for future research.
}

We consider group sizes of 1 (individual), 5, 10, and 15, creating
five groups for each size, totaling 20 human agents spanning 155 distinct players.
Our human participants, all fluent in US English, are experienced Quiz Bowl players.
While this sample may not encompass the full diversity of the broader population,
their expertise in trivia games, particularly in Quiz Bowl,
allows us to contrast the nuanced skill sets of seasoned Quiz Bowl enthusiasts
with the capabilities of our \ai{} systems.
%

\subsection{\ai{} Agents}\label{subsec:ai-agents}
To capture skill differentials across \ai{} models and humans and to learn the effects of
various training and modeling techniques, we select a broad range of \qa{}
systems,\footnote{\autoref{appendix:qa-models} provides further details into model specs.}
grouped as below:

\paragraph{Retrievers.}
These agents, indexing Wikipedia, use sparse (e.g., \bmtfive{}), and
dense---\abr{grit-lm}~\cite{muennighoff2024generative} and
\abr{contriever}~\cite{izacard2021unsupervised}---methods
to fetch the $k$ most relevant context documents to a query (where $k$ = 1, 3, 5, 10).
We call these context-retrievers.
We also test a title-retriever, where only the title(s) associated with the retrieved document(s)
are answer predictions.
Retrievers are evaluated on recall, with a point scored if the answer
appears within retrieved documents for context-retrievers, or in the title for the title-retrievers.

\paragraph{Large Language Models (\llm{s}).}
We assess \llm{s} zero-shot in-context learning~\cite{Brown2020LanguageMA},
providing a task instruction followed by a single \qa{} pair demonstration.
These \llm{s} include \emph{base} models
(OPT~\cite{Zhang2022OPTOP}, GPT-Neo~\cite{Black2021GPTNeoLS} and Pythia~\citep{biderman2023pythia}),
instruction-tuned models (OPT-IML~\cite{iyer2022optiml},
T0, T0pp~\cite{sanh2021multitask}, Flan-T5~\cite{Chung2022ScalingIL} and
Flan-UL2~\cite{tay2022ul2}),
very large-scaled models like \llama{}~\cite{touvron2023llama}, Falcon40B~\cite{falcon40b},
Cohere's \abr{cmd-r+}~\footnote{https://docs.cohere.com/docs/command-r-plus} and Mixtral 8x7b~\cite{jiang2024mixtral}, and
closed-sourced APIs such as \gptfouro{}, \gptfour{}~\cite{openai2023gpt4} and Gemini-family~\cite{team2024gemini}.
%

\paragraph{Retriever-augmented Generative Models (\abr{rag}).}
%
We combine above defined retrievers with generative models
for answer production, primarily using FlanT5-XL~\cite{Chung2022ScalingIL} with top 3
documents and exploring Flan-UL2~\cite{tay2022ul2}, and \abr{cmd-r+} to accommodate all ten.

\paragraph{Answer Match Equivalence.}
Traditional exact-match~\citep{rajpurkar-16} often misses alternative answer
that have different wordings or forms but the same semantic sense as the correct
answer~\cite{bulian2022tomayto}.
%
%
To better handle this, we adopt a fuzzy match evaluation using answer aliases~\cite{Si2021WhatsIA}: if the character level 
matching rate between the predicted answer and the gold answer exceeds a
certain threshold, the prediction is considered as correct.
We tuned the threshold against human judgments on a small dev set.

\subsection{\caimira{} Setup}\label{subsec:kira-setup}
\begin{figure}[t]
  \centering
  \includegraphics[width=\linewidth]{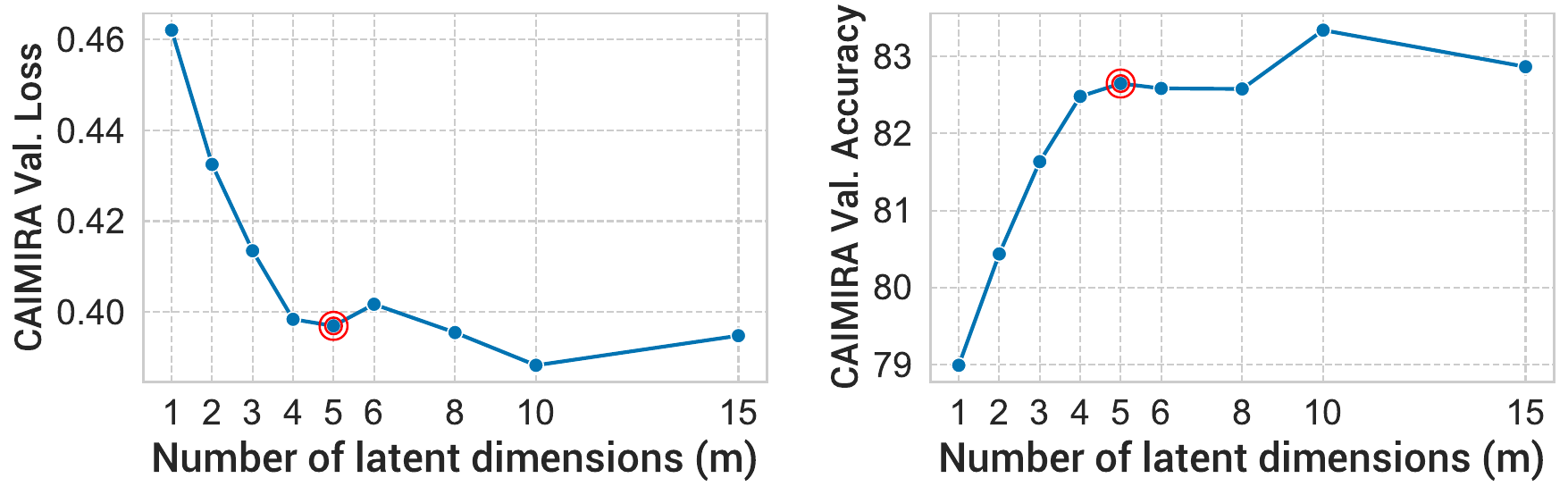}
  \caption{Ablation study showing \caimira{} performance with varying latent dimensions $m$, indicating sufficiency at $m=5$, beyond which gains are marginal.}\label{fig:dim-ablation}
\end{figure}

We ablate the number of latent dimensions, $m$.
Validation loss plateaus beyond $m=5$ (Fig~\ref{fig:dim-ablation}).
%
We thus train a $5$-dimensional \caimira{} model using \texttt{\small{all-mpnet-base-v2}},
an \abr{sbert} variant~\citep{reimers2019sentencebert} as the question embedder $f_E$.
To capture information gaps between questions and answers,
we supplement \abr{sbert}'s text input with both the answer and it's Wikipedia page summary.
We minimize $\mathcal{L}_{\caimira}$~(\autoref{eq:loss-caimira}) using Adam optimizer~\citep{kingma2014adam},
with learning rate 0.005, batch size $512$, and $\lambda_d = \lambda_s = 1e-5$.

\begin{figure*}[th]
  \centering
  \includegraphics[width=\linewidth]{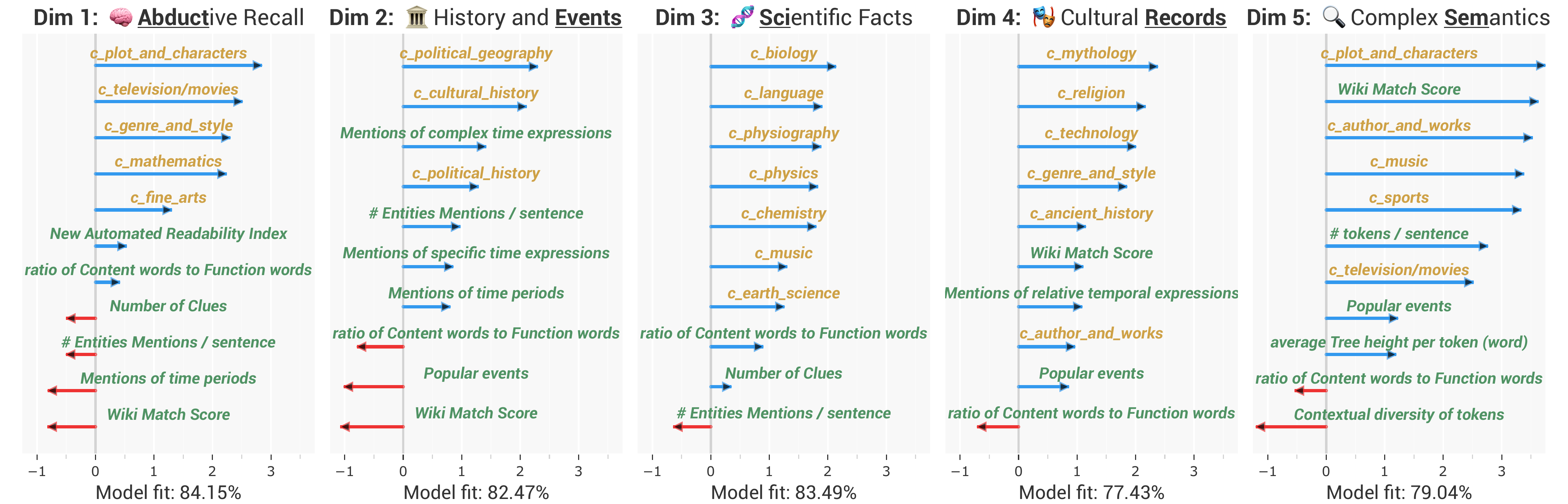}
  \caption{\textbf{Interpretation of the five latent dimensions in \caimira{}}.
    We use Logistic Regression to predict the binary relevance label,
    $\rel{jk} > 0.6$, for each dimension $k$. For question features, we use
    \textcolor{caimiradarkyellow}{\textbf{topical categories}} and
    \textcolor{caimiradarkgreen}{\textbf{linguistic properties}}. We report the
    classification accuracy and the statistically significant features.
    Coefficients are \textcolor{dullblue}{\textbf{positive}} if the features positively affect
    classification, \textcolor{dullred}{\textbf{negative}} otherwise. This demonstrates the efficacy
  of predicting the relevance from a question's \abr{sbert} embedding.}\label{fig:relevance-analysis}
\end{figure*}

\begin{figure*}[ht]
  \centering
  \includegraphics[width=\linewidth]{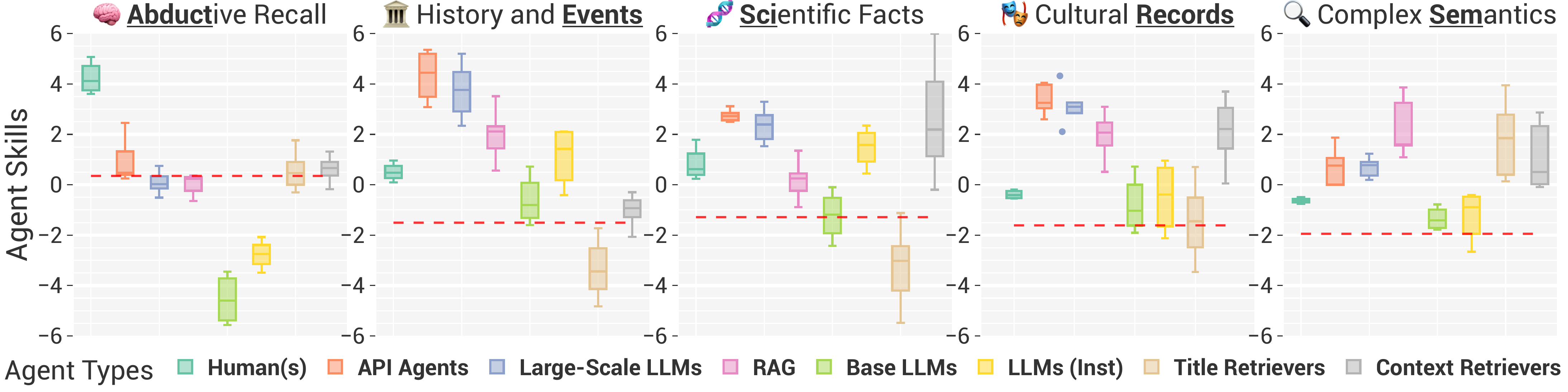}
  \caption{Distribution of skills grouped by agent type across the five latent dimensions of \caimira{}.
  Interpretations given in~\autoref{fig:relevance-analysis}.
  The red dashed line indicates the mean \emph{effective difficulty} of each dimension~(\autoref{eq:mean-effective-difficulty}).}\label{fig:agent-skills}
\end{figure*}

\paragraph{Interpreting Latent Aspects.}\label{subsec:logreg-setup}
To study the latent dimensions of \caimira{}, we use Logistic Regression
as a supplemental interpretative tool.
%
We build upon~\citet{benedetto2020r2de},
which uses Linear Regression to post-hoc explain the latent item difficulty parameters,
and follow~\citet{gor-etal-2021-toward} to interpret the latent relevance dimensions using logistic regression.
For each latent dimension ($k$), Logistic Regression predicts if the relevance $\rel{jk}$
is greater than $0.6$ as a function of interpretable features extracted from the questions.
These features span topical question subcategories, clue counts,
temporal expression mentions, question similarity with
corresponding Wikipedia pages (WikiMatchScore), and
linguistic features from \citet{lee-etal-2021-pushing}.\footnote{~\autoref{appendix:logreg-feats} lists all features we use.}
Thereby, we explain \caimira{}'s latent dimensions by relating them to the
logistic regression features with large (positive and negative) coefficients.
Topical features are one-hot encoded; \feat{c\_music} is set to 1
for music related question, and 0 otherwise.
The linguistics features span advanced semantic, discourse-based, and
syntactic elements, providing a rich and multi-faceted representation
of the questions. These are normalized to have zero mean and unit variance.
\autoref{fig:relevance-analysis} lists the most contributing, statistically significant
features for each dimension ($p\text{-value} < 0.05$).
To make the learned coefficients comparable across dimensions,
we incorporate class-balancing maintaining the random guess accuracy for each
dimension at 50\%.

\section{Question and Agent Analysis}\label{sec:qanda}
\newcommand{\dimone}{\underline{Abduct}ive Recall}
\newcommand{\dimtwo}{History and \underline{Events}}
\newcommand{\dimthree}{\underline{Sci}entific Facts}
\newcommand{\dimfour}{Cultural \underline{Records}}
\newcommand{\dimfive}{Complex \underline{Sem}antics}

This section interprets the latent aspects of \caimira{},
emphasizing their role in differentiating agent skills.
It also examines the patterns of question difficulty and agent performance.

\subsection{Latent aspects and Agent skills}\label{subsec:relevance-analysis}
\caimira{} uncovers five latent aspects, each capturing distinct question styles and
content, determined by specific linguistic and topical features (\autoref{fig:relevance-analysis}).
These aspects highlight varying agent skills across the latent dimensions (\autoref{fig:agent-skills}).
In naming and interpreting these aspects, we draw on educational assessment frameworks,
particularly Bloom's Taxonomy~\citep{anderson2001taxonomy}, which emphasizes the
stages of knowledge recall, comprehension, and application---skills central to the Quizbowl dataset.

\newcommand{\quizquestion}[1]{\textvtt{{\small"#1"}}}
\paragraph{\dimone.}
The first aspect captures a cognitive process that combines elements of
inferential reasoning with targeted knowledge retrieval.
It requires bridging indirect clues and vague references to formulate the information gap,
and recalling specific entities to fill the gap.
This distinguishes it from purely creative and commonsense-based
abductive reasoning tasks in linguistics literature~\cite{bhagavatula2019abductive,shi2024language}.
We term this aspect ``abductive recall'' to highlight the interplay between
hypothesis generation and gap resolution through targeted fact retrieval.
Questions often narrate events and describe characters from a fictional realm
while deliberately avoiding direct references to named entities or key phrases~(Example in Fig~\ref{fig:workflow}).
A low WikiMatchScore---semantic overlap between questions and their associated Wikipedia
pages---combined with the absence of entities and key phrases, indicate a significant
information gap that necessitates not just multi-hop reasoning skills to bridge the
contextual divide, but also deducing relevant involved entities from the narrative.
Humans excel at these questions, surpassing \gptfour{} by leveraging intuition to
connect abstract clues to specific entities, while most \abr{ai} models struggle.

\paragraph{\dimtwo.}
In contrast, the second dimension involves historically grounded questions,
where the information gap is clearer, though the queries are complex.
These questions challenge participants to synthesize multiple pieces of information
and infer connections between events.
For e.g, \quizquestion{This man
was killed by a crossbow bolt while besieging the castle Charlus-Chabrol},
requires identifying both the event and the historical figure.
While these questions still feature lower WikiMatchScores, the gap is more structured,
centering around entity relations like events, people, and places.
Bigger \llm{s} excel in this category, often outperforming humans and retrievers,
suggesting effective recall and application of historical information through their parametric memory.

\begin{figure}[h!]
  \centering
  \includegraphics[width=\linewidth]{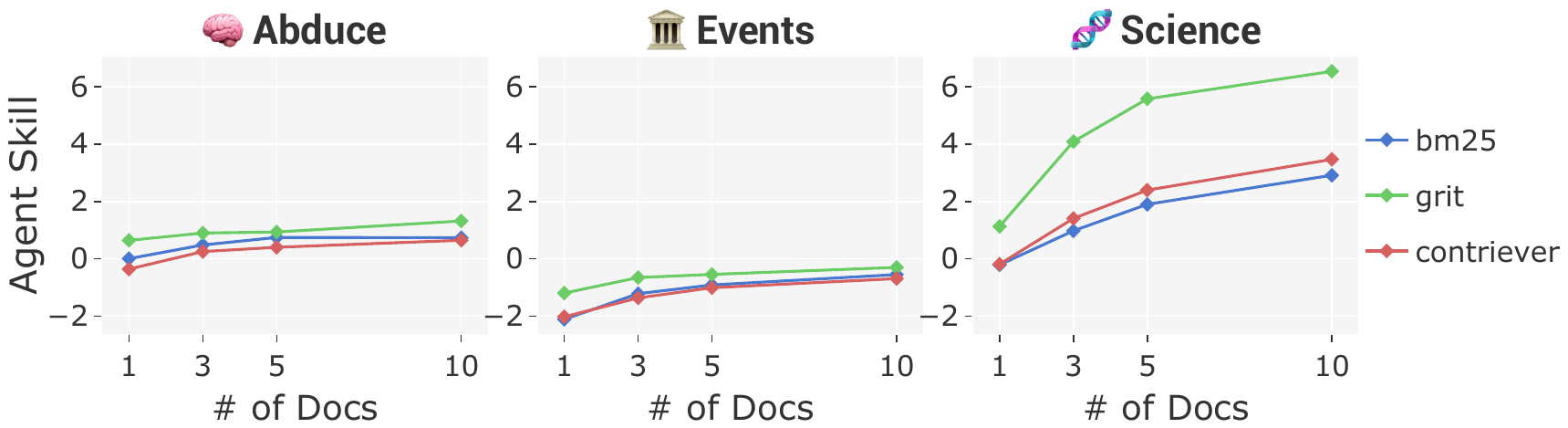}
  \caption{Variation in Context Retriever skills across latent dimensions as the number of retrieved documents
    (top-$k$) increases, showing that a system which retrieves more documents can achieve higher skills in \emph{Science}, but not on \emph{Abduction} and \emph{Events}.}
  \label{fig:retriever-skills}
\end{figure}

\paragraph{\dimthree.}
This aspect focuses on domain-specific conceptual knowledge,
often featuring questions from scientific domains.
Retrieval-based systems fare well when allowed to retrieve sufficient documents~(\autoref{fig:retriever-skills}).
Notably, these questions, along with history-related ones, best differentiate
instruction-tuned \llm{s} from base models, with the former
outperforming the latter. Humans and large-scale \llm{s} excel in this category,
as do closed-source systems like \gptfour{}.

%

\begin{figure}[t!]
  \centering
  \includegraphics[width=\linewidth]{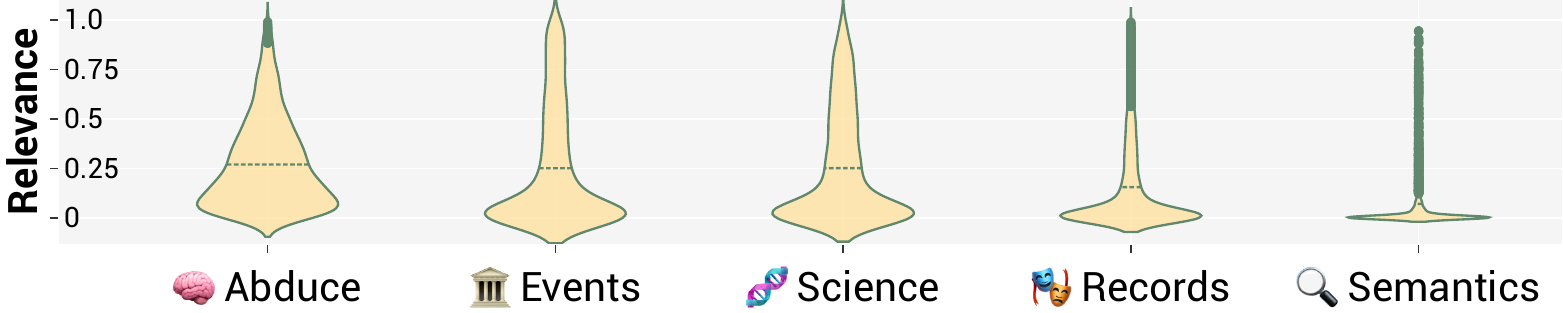}
  \caption{Distribution of \emph{relevance} ($r_{j,k}$) scores across \caimira's five latent dimensions.
  \dimfour{} and \dimfive{} are not as representative of the dataset, as the first three.}
  \label{fig:relevance-distribution}
\end{figure}
\vspace{-3mm}

\paragraph{\dimfour.}
This aspect represents questions focusing on prominent figures such as
authors, composers, artists, and leaders, asked in the style of ``who did what'',
testing direct knowledge recall of well-known facts and making them relatively easy
and accessible (high WikiMatchScore).

\paragraph{\dimfive.}
The final aspect pertains to questions about popular events,
featuring complex semantic relationships and detailed sentences with less common,
domain-specific keywords.
Despite their intricacy, they are particularly retriever-friendly due to high
WikiMatchScores, indicating a significant overlap with relevant source documents.
The most prominent fact about the answer is directly mentioned in both the question
and the document, enabling retrievers to locate correct documents.
However, agents without retrieval abilities, or large parametric memories, struggle.

\begin{figure}[h!]
  \centering
  \includegraphics[width=\linewidth]{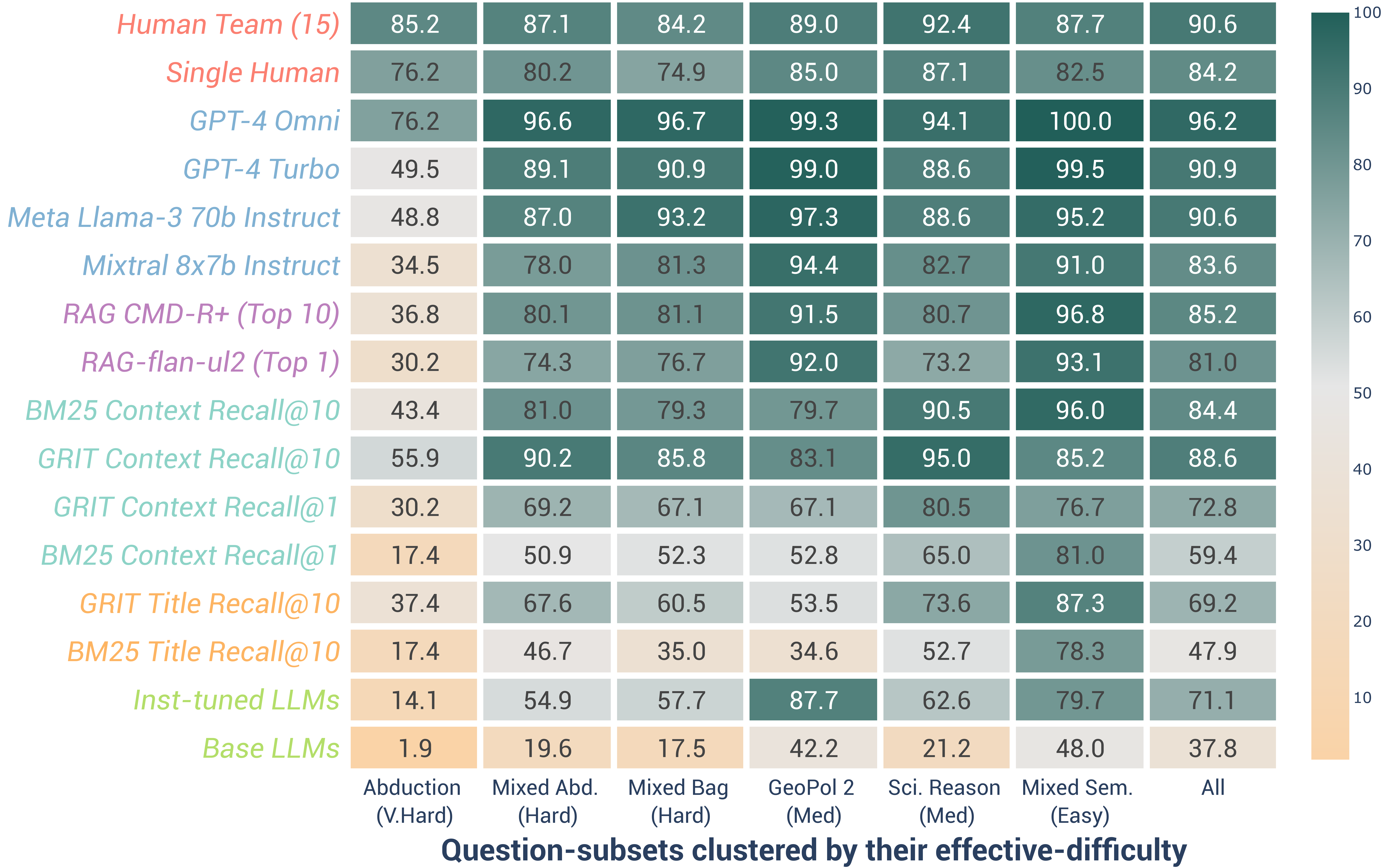}
  \caption{Agent accuracies on various dataset slices.}
  \label{fig:agent-accuracies}
\end{figure}
\vspace{-5mm}
%
%
\subsection{Which Questions are most difficult?}\label{subsec:diff-analysis}
%
To identify groups of questions that present different challenges, we analyze each question’s
\emph{effective difficulty}, denoted as $\ediff{j,k}$. This metric represents the
contribution of the $k$-th latent aspect to the difficulty of question $j$,
calculated as $\rel{j,k}\diff{j,k}$ according to Equation~\ref{equ:kira}.
%
%
We cluster questions into twelve groups using KMeans on their 5-dimensional effective difficulty
$\ediff{j}$, then analyze \emph{mean relevance} and \emph{mean effective difficulty} per cluster
across dimensions (Fig~\ref{fig:effective-difficulty}, full set in \autoref{appendix:question-difficulty}).
The mean effective difficulty $\ediff{D, \mu_k}$ on the dimension $k$ for a question
set $D$ is calculated as a weighted mean of the effective difficulty scores over
the questions in $D$, normalized by the total relevance.
\begin{align}
  \ediff{D, \mu_k} = \frac{\sum_{j\in D}\rel{j,k}\diff{j,k}}{\sum_{j\in D}\rel{j,k}} \label{eq:mean-effective-difficulty}
\end{align}
%
%
\emph{Abduction (V.Hard)} and \emph{Mixed Bag} emerge as the most challenging categories,
demonstrating high difficulty due to complex semantics, indirect phrasing and also
mostly having a single clue. \ai{} systems, including \gptfour{}, struggle with these,
highlighting a marked disparity with human accuracy~(Fig~\ref{fig:agent-accuracies}).
%
Instruction-tuned \llm{s} outperform base ones in moderately difficult science
questions, with \gptfouro{} surpassing single human players.
A common trend we observe is that for each latent factor, questions tend to have higher difficulty
when they have fewer clues, and lower WikiMatchScore.
\begin{figure}[t!]
  \centering
  \includegraphics[width=\linewidth]{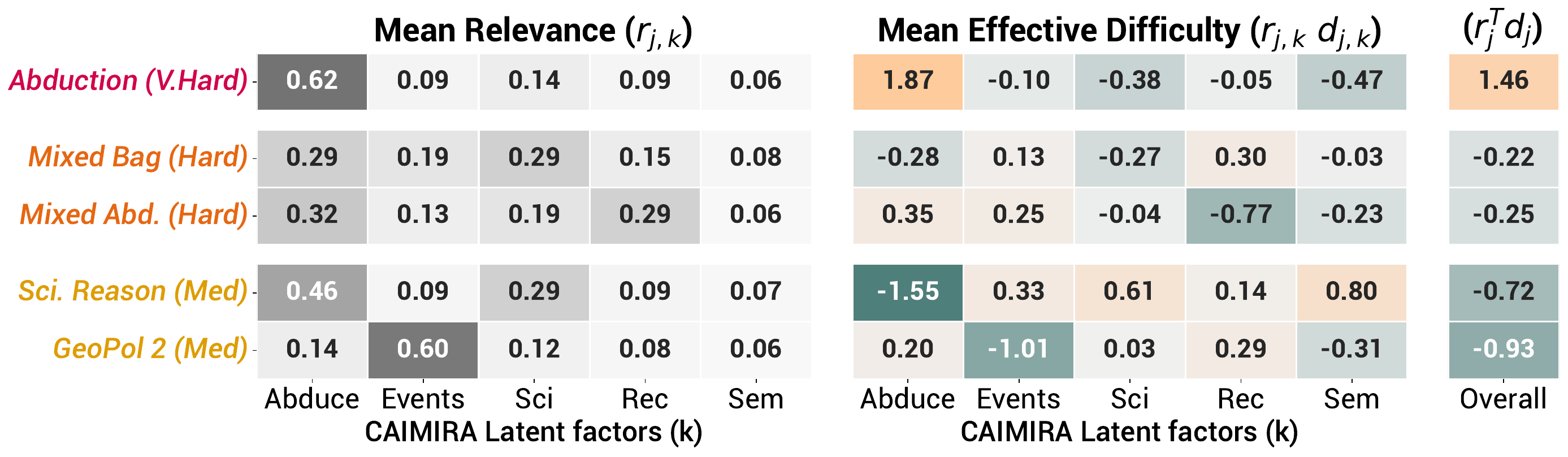}
  \caption{Heatmaps of mean relevance $\rel{j,k}$ and \emph{mean effective difficulty} $\ediff{D, \mu_k}$
  of selected question clusters (on effective difficulty) across the five latent factors ($k$).
  }\label{fig:effective-difficulty}
\end{figure}

\section{Related Work}
\paragraph{Adoption of IRT in NLP.}\label{subsec:irt-adoption}
%
Current evaluation paradigms for machine and human \qa{} inadequately segment datasets,
treating questions as independent single transaction without assessing relative
differences between the test set items.
To remedy this, \citet{lalor2019learning} propose adopting the \irt{} ranking method
from educational testing as a novel evaluation framework for \nlp{}.
\citet{rodriguez2021evaluation} argue for the adoption of \irt{} as the de facto
standard for \qa{} benchmarks, demonstrating its utility in guiding annotation effort,
detecting annotator error, and revealing natural partitions in evaluation datasets.
\citet{byrd-srivastava-2022-predicting} further uses \irt{} to estimate question
difficulty and model skills, and use question features to post-hoc predict question difficulty.
Yet, existing studies are confined to a one-dimensional \irt{} models. Our research
advances this domain by enhancing the learning method and capturing question traits
that effectively differentiate human and \ai{} \qa{} abilities.


\paragraph{Ideal Point Models (IDP)}\label{subsec:idp}
\irt{} and \abr{ipm} are two prominent statistical models used in different fields for
distinct purposes. 
Both models deal with the analysis of preferences or abilities,
but their applications and theoretical underpinnings show significant differences.
\irt{}, used in educational assessments, gauges abilities from question responses,
typically focusing on one-dimensional traits~~\cite{de2013theory}.
Conversely, \abr{ipm}, applied in political science, evaluates positions on spectra
like political ideologies based on choices or votes~\cite{clinton2004statistical}.
Despite differences, both employ mathematically equivalent probabilistic methods
to estimate the likelihood of a binary outcome---correctness in \irt{}, and
votes in \abr{idp}, from a set of covariates, such as question difficulty
or political ideology.
%

\paragraph{Human-AI Complementarity.}\label{subsec:complement}
Research in \nlp{} has increasingly focused on augmenting human skills with language models,
particularly in the areas like creative writing and question-answering.
Studies have explored collaborative writing with \llm{s}, such as having human writers
use \gptthree{} for suggestions~\citep{Lee2022CoAuthorDA} or modifying user-selected text
spans for enhanced descriptiveness~\citep{Padmakumar2021MachineintheLoopRF}.
For trivia, experts and novices have teamed up with \ai{}~\citep{Feng2018WhatCA},
and for information retrieval, humans used \ai{}-generated queries to find
answers~\citep{He:Mao:Boyd-Graber-2022}
Our approach diverges by focusing modeling latent factors that best 
accentuate the distinct capabilities of trivia nerds and \ai{} in \qa{}.
This strategy aims to identify the benchmarking methods for assessing and enhancing
\ai{} systems in subsequent work.

\jbgcomment{I'd focus more on adversarial examples, BERTology probes of what models are capable of, etc.  }

\section{Conclusions}
\vspace{-2mm}
\caimira{} enables discovery and interpretation of latent aspects in \qa{} datasets
that highlight the skills of various \qa{} agents.
On contrasting \abr{ai} systems with humans, we find notable disparities: systems like
\gptfour{} and Gemini Pro excel at direct, context-rich queries that require connecting events and figures,
but struggle with indirectly phrased questions lacking explicit entity references---domains where human acumen shines.
Although \gptfour{} matches individual human performance on complex knowledge-intensive
abductive reasoning tasks, we caution against interpreting this as indicative of
superhuman abilities.
Given that the quiz questions that Protobowl is based off have been publicly available
since 2011, and that these models' training data is not fully known, accurately assessing
the reason for their near-perfect performance is challenging.
Future research should aim to develop stronger and innovative evaluations that better
gauge \ai{} systems' ability to understand implicit contexts, and systematically contrast
their skills with those of humans.
Lastly, this work opens up new avenues for research on estimating agent skills that can
be combined to assess multi-agent systems and collaborations, which becomes crucial as
\abr{nlp} evolves toward conversational agents and real-world problem-solving.\label{sec:conclusion}


\section{Limitations}
\paragraph{Dataset and Task Limitations}
Our study faces constraints related to dataset and task setup:
(1) Limited language diversity: Our English-only dataset restricts generalizability to other languages.
(2) Lack of diverse task types: We rely solely on trivia-based questions, lacking non-trivia datasets with human responses in competitive settings.
(3) Absence of multilingual trivia benchmarks: We lack multilingual trivia datasets with human responses and performance benchmarks.
Future work should address these by creating datasets that include non-trivia tasks,
multiple languages, and human responses, offering a more comprehensive understanding
of human and AI performance across diverse linguistic and task environments.

\paragraph{Challenges in interpreting near-perfect scores}
While models like \gptfour{} match or exceed individual humans on complex tasks,
caution is needed when interpreting these results as superhuman. Quiz questions in our
Protobowl-based dataset have been public since 2011, and the models' full training data
is unknown. This makes it difficult to determine if their near-perfect performance
stems from genuine reasoning or exposure to specific questions during pre-training.
genuine reasoning or exposure to specific questions during pre-training. This
limitation highlights the need for more robust evaluation methods to accurately
assess AI systems' understanding and reasoning abilities compared to humans.

\paragraph{Lack of information on specific human players}
Because of the nature of the Protobowl platform that we used to collect the human response
data, we do not have access to information about the specific human players to incorporate
that into our analysis. Future work can focus on collecting such information
whilst hiding the user identity.

\paragraph{Non-extensibliity of a trained \caimira{} to a new \ai{} systems.}
Unlike how \caimira{} extended \mirt{} to model question characteristics as a function of
question texts, and not just unique question identifiers, \caimira{} is not extensible to
a new agent without retraining the model. To make this possible for \ai{} systems,
future work can maintain a feature set that describes the specifications of an \ai{}
system that can include the model architecture, the training data, parameters, training
strategies, etc, and have \caimira{} learn a transformation from the feature set to agent skills.
However, since this approach would require having a feature set for human players as well,
which is not available, this approach is not feasible at the moment.

\paragraph{Static representation from \abr{sbert}.}
In this work, we use a static dense representation of the question text from \abr{sbert},
instead of finetuning the model for adapting to \caimira{} objective that learns
representations from question text that best predicts the human response.
This was out of the scope of this study. Future work can explore this direction using
parameter efficient finetuning (\abr{PEFT})~\cite{xu2023parameterefficient}.


\label{sec:limitations}

\section{Ethical Considerations}
In conducting this study, we adhered to strict ethical guidelines
to ensure respect for privacy, obtaining informed consent from
human participants and annonimization of their data.
Our work complies with all relevant ethical standards,
underscoring our commitment to ethical research practices in
advancing NLP technologies.
We utilized GitHub Copilot for low level coding and writing assistance---reimplementing plotting codes,
as well as editing the prose in this document to improve readability and conciseness.

Regarding ethical considerations about running computationally expensive models, we
acknowledge the carbon footprint of training and running large-scale language models.
In our study we only train a very small of order 25000 parameters, for 20 minutes
of single A4000 GPU time. We also use a pre-trained \abr{SBERT} model for encoding the question text.\label{sec:ethics}

\section{Acknowledgments}
We thank the University of Maryland's \abr{clip} lab members: Neha Srikanth,
Navita Goyal, Rupak Sarkar, along with the alumni: Pedro Rodriguez, Sweta Agrawal,
and Chenglei Si for useful discussions and valuable feedback. We also thank John
Kirchenbauer for his suggestions on the toolings used for experimental evaluations.
We thank Ryan Rosenberg and Ophir Lifshitz for their discussions of buzzpoint data.
This material is based upon work supported by the National Science
Foundation under Grant No. \abr{iis}-2403436 (Boyd-Graber) and the Army Research
Office under Grant Number W911NF-23-1-0013 (Gor). 
Any opinions, findings, views, conclusions, or recommendations expressed
in this material are those of the author(s) and do not necessarily reflect
the views of the National Science Foundation or the official policies of
the Army Research Office or the U.S. Government.
The U.S. Government is authorized to reproduce and distribute reprints for
Government purposes notwithstanding any copyright notation herein. 
Finally, we express our gratitude to Flaticons\footnote{\url{https://www.flaticon.com/}}
for their extensive collection of icons which we utilize for making figures in this work.
\label{sec:ack}

\bibliography{bib/jbg, bib/custom}

\clearpage
\appendix
\section{Quizbowl Dataset}\label{appendix:qb}
Quizbowl~\citep{rodriguez2019quizbowl}, the source of questions for \pb, is a trivia game consisting of questions with clues decreasing in difficulty and culminating with a "giveaway" hint at the end of the question. The sequence of clues often reveals more information or helps disambiguate possible references and interpretations at each step. \autoref{fig:appendix-qb-example} illustrates this structure with three example questions from different categories.

\begin{figure}[ht]
    \tiny
    \begin{tabularx}{\linewidth}{X}
        \textvtt{{\textcolor{qbred}{Question ID\: q832\_5 (Category: Religion)}}} \\

        \textvtt{\textcolor{qbred}{This text was written down by Sahabas (sah-HAH-bahs) after the death of the leader that received it. The clarification of the meaning and significance of this document is the practice of tafsir (TAHFSEER). Its hundred and fourteen chapters are called suras (soor-AHS). It literally means "the recitation" and is said to have been revealed by Gabriel to Muhammad. For 10 points, what "divinely ordained" religious text is sacred to Muslims?}} \\
        \textvtt{\textcolor{qbred}{\underline{Answer: Piano / Pianoforte}}} \\ \midrule

        \\ \textvtt{{\textcolor{qbred}{\textbf{Question ID\: q622\_3} (Category: Music)}}} \\
        
        \textvtt{\textcolor{qbred}{Paul Wittgenstein commissioned concertos for this instrument that used only the left hand. This instrument is said to have been invented by Bartolomeo Cristofori ("BAR-tow-lo- MAY-oh KRIS-tow-for-ee"). It was originally named for its ability to play both loud and soft sounds, which made it an improvement over the clavichord and harpsichord.}} \\
        \textvtt{\textcolor{qbred}{\underline{Answer: Piano / Pianoforte}}} \\ \midrule

        \\ \textvtt{{\textcolor{qbred}{\textbf{Question ID\: q2443\_1} (Category: Science > Mathematics)}}} \\
        
        \textvtt{\textcolor{qbred}{4 times the infinite sum one, minus one third, plus one fifth, minus one seventh, et cetera, equals this number.}} \\
        \textvtt{\textcolor{qbred}{\underline{Answer: pi / 3.14 / $\pi$}}}
    \end{tabularx}
    \caption{Example of QuizBowl questions for three different categories: Religion, Music and Mathematics, that illustrates the incremental nature of the questions.}\label{fig:appendix-qb-example}
\end{figure}

Quizbowl naturally discriminates players' skills as players can \textbf{interrupt} questions to answer, and answering earlier is better.

In contrast to ``all or nothing'' \qa{}, incremental \qb{} questions help
pinpoint the clues necessary for an agent $a$ to answer question $q$ by creating multiple
opportunities for $a$ to answer $q$.
We achieve this by creating creating multiple entries for a single quizbowl question into our dataset.
For instance, if a Quizbowl question \qbid{q622} has four clues in total,
we create four entries, viz. \qbid{q622\_1}, \qbid{q622\_2}, \qbid{q622\_3}, and \qbid{q622\_4}, each
corresponding to the question with first $i$ clues, where $i \in \{1, 2, 3, 4\}$.
%

\section{\caimira{} Setup.}
In this section, we provide a detailed explanation of the learning objective for \caimira{} and the hyperparameters used in our experiments.
First, let's revise the \caimira{} objective from Section~\ref{sec:kira}:
\begin{align}
    &p(U_{i, j} = 1 \g \skill{i}, \rel{j}, \diff{j}) = \sigma\left({(\skill{i} -  \diff{j})}^{\intercal} \rel{j}\right). \nonumber \\
    &\text{\footnotesize where, $\skill{i} \in \mathbb R^m$ is agent skills, } \nonumber \\
    &\text{\footnotesize and, $\rel{j}, \diff{j}\in\mathbb R^m$ are question relevance and difficulty resp.} \nonumber
\end{align}
Here, $\diff{i}$ and $\rel{j}$ are functions of question representation $\mathbf{E}^q_j$ defined as:
\begin{align}
\mathbf{r'_j} &= \mathbf{W}_R\ \mathbf{E}^q_j + \mathbf{b}_R, & \mathbf{d'_j} &= \mathbf{W}_D\ \mathbf{E}^q_j, \nonumber \\
\rel{j} &= \text{softmax}(\mathbf{r'_j}), & \diff{j} &= \mathbf{d'_j} - \frac{1}{n_q}\sum_{j=1}^{n_q} \mathbf{d'_j}, \nonumber
\end{align}
where $\mathbf{W}_R, \mathbf{W}_D \in \mathbb{R}^{m \times n}$ and
$\mathbf{b}_R \in \mathbb{R}^{m}$. 
These, along with the embedding matrix
$\mathbf{E}^a$ of agent skills~($\skill{i} = \mathbf{E}^a_i$), are the
parameters we train for \caimira{} over a regularized cross entropy objective.

\paragraph*{Hyperparameters.}
The trainable parameters are fit using mini-batch stochastic gradient descent to
minimize $\mathcal{L}_{\caimira}$~(\autoref{eq:loss-caimira}),
where $\lambda_d$ and $\lambda_s$ are set to $1e-5$.
We use Adam optimizer~\citep{kingma2014adam} without weight decay, and with a 
learning rate of $0.005$, and the batch size is set to $512$.
\section{\qa{} Agents in our study}\label{appendix:qa-models}
This section describes the \qa{} agents used in our study, including the retrievers, \llm{s}, \abr{rag} models, 
and the prompts used to query them.
\begin{figure}[h]
        \centering
    
        \begin{tcolorbox}[colback=agentblue!5!white,colframe=agentblue!75!black,title=Contexts Recall@10, fonttitle=\bfseries\tiny\ttfamily, coltitle=white, fontupper=\tiny\ttfamily]
			\textbf{bm25\_ctx-recall@10}  \\

			\textbf{contriever\_ctx-recall@10}
        \end{tcolorbox}
        
        \begin{tcolorbox}[colback=agentblue!5!white,colframe=agentblue!75!black,title=Contexts Recall@3, fonttitle=\bfseries\tiny\ttfamily, coltitle=white, fontupper=\tiny\ttfamily]
			\textbf{bm25\_ctx-recall@3}  \\

			\textbf{contriever\_ctx-recall@3}
        \end{tcolorbox}
        
        \begin{tcolorbox}[colback=agentblue!5!white,colframe=agentblue!75!black,title=Top Context, fonttitle=\bfseries\tiny\ttfamily, coltitle=white, fontupper=\tiny\ttfamily]
			\textbf{bm25\_ctx-recall@1}  \\

			\textbf{contriever\_ctx-recall@1}
        \end{tcolorbox}
        
    \caption{Agents we use in the Context Retrievers category.}\label{fig:appendix-agent-names-context-retrievers}
\end{figure}

\begin{figure}[h]
        \centering
    
        \begin{tcolorbox}[colback=agentlightblue!5!white,colframe=agentlightblue!75!black,title=Title Recall@10, fonttitle=\bfseries\tiny\ttfamily, coltitle=white, fontupper=\tiny\ttfamily]
			\textbf{bm25\_title-recall@10}  \\

			\textbf{contriever\_title-recall@10}
        \end{tcolorbox}
        
        \begin{tcolorbox}[colback=agentlightblue!5!white,colframe=agentlightblue!75!black,title=Title Recall@3, fonttitle=\bfseries\tiny\ttfamily, coltitle=white, fontupper=\tiny\ttfamily]
			\textbf{bm25\_title-recall@3}  \\

			\textbf{contriever\_title-recall@3}
        \end{tcolorbox}
        
        \begin{tcolorbox}[colback=agentlightblue!5!white,colframe=agentlightblue!75!black,title=Top Title, fonttitle=\bfseries\tiny\ttfamily, coltitle=white, fontupper=\tiny\ttfamily]
			\textbf{bm25\_title-recall@1}  \\

			\textbf{contriever\_title-recall@1}
        \end{tcolorbox}
        
        \begin{tcolorbox}[colback=agentlightblue!5!white,colframe=agentlightblue!75!black,title=Inst Title Retriever R@10, fonttitle=\bfseries\tiny\ttfamily, coltitle=white, fontupper=\tiny\ttfamily]
			\textbf{grit\_title-recall@10}
        \end{tcolorbox}
        
        \begin{tcolorbox}[colback=agentlightblue!5!white,colframe=agentlightblue!75!black,title=Inst Title Retriever R@3, fonttitle=\bfseries\tiny\ttfamily, coltitle=white, fontupper=\tiny\ttfamily]
			\textbf{grit\_title-recall@3}
        \end{tcolorbox}
        
        \begin{tcolorbox}[colback=agentlightblue!5!white,colframe=agentlightblue!75!black,title=Inst Title Retriever R@1, fonttitle=\bfseries\tiny\ttfamily, coltitle=white, fontupper=\tiny\ttfamily]
			\textbf{grit\_title-recall@1}
        \end{tcolorbox}
        
    \caption{Agents we use in the Title Retrievers category.}\label{fig:appendix-agent-names-title-retrievers}
\end{figure}

\paragraph{Retrievers as \qa{} agents.} Our retrievers, which index Wikipedia documents,
respond with the top $k$ documents (where $k$ = 1, 3, 10) most relevant to the question.
We employ two types of retrievers: dense and sparse. 
The dense retriever, \abr{contriever}~\cite{izacard2021unsupervised}, is pretrained via
unsupervised contrastive learning on a mix of Wikipedia and CCNet data and 
then fine-tuned on MS-MARCO~\cite{campos2016ms}.
The sparse retriever utilizes the \abr{BM25} algorithm~\cite{10.1561/1500000019} and
Anserini's implementation with index~\cite{Lin_etal_SIGIR2021_Pyserini}.
We also test a title-retriever, assuming the document title is the query answer.
Retrievers are evaluated on recall-based accuracy, with a point scored if the answer
appears within the top-$k$ documents for context-retrievers,
or in the title of the top-$k$ documents for the title-retriever.

\begin{figure}[ht]
        \centering
    
        \begin{tcolorbox}[colback=agentpink!5!white,colframe=agentpink!75!black,title=40b+ LLMs, fonttitle=\bfseries\tiny\ttfamily, coltitle=white, fontupper=\tiny\ttfamily]
			\textbf{cohere-command-r-plus\_1shot}  \\

			\textbf{falcon-40b-instruct\_1shot}  \\

			\textbf{falcon-40b\_1shot}  \\

			\textbf{llama-2-70b\_1shot}  \\

			\textbf{meta-llama-3-70b-instruct\_1shot}  \\

			\textbf{meta-llama-3-70b\_1shot}  \\

			\textbf{mixtral-8x7b-instruct\_1shot}
        \end{tcolorbox}
        
        \begin{tcolorbox}[colback=agentpink!5!white,colframe=agentpink!75!black,title=Inst Ctx Retriever R@10, fonttitle=\bfseries\tiny\ttfamily, coltitle=white, fontupper=\tiny\ttfamily]
			\textbf{grit\_ctx-recall@10}
        \end{tcolorbox}
        
        \begin{tcolorbox}[colback=agentpink!5!white,colframe=agentpink!75!black,title=Inst Ctx Retriever R@3, fonttitle=\bfseries\tiny\ttfamily, coltitle=white, fontupper=\tiny\ttfamily]
			\textbf{grit\_ctx-recall@3}
        \end{tcolorbox}
        
        \begin{tcolorbox}[colback=agentpink!5!white,colframe=agentpink!75!black,title=Inst Ctx Retriever R@1, fonttitle=\bfseries\tiny\ttfamily, coltitle=white, fontupper=\tiny\ttfamily]
			\textbf{grit\_ctx-recall@1}
        \end{tcolorbox}
        
        \begin{tcolorbox}[colback=agentpink!5!white,colframe=agentpink!75!black,title=Base LLMs, fonttitle=\bfseries\tiny\ttfamily, coltitle=white, fontupper=\tiny\ttfamily]
			\textbf{gpt-neo-2.7B\_1shot}  \\

			\textbf{opt-2.7b\_1shot}  \\

			\textbf{pythia-12b-deduped\_1shot}  \\

			\textbf{pythia-12b\_1shot}  \\

			\textbf{pythia-2.8b-deduped\_1shot}  \\

			\textbf{pythia-2.8b\_1shot}  \\

			\textbf{pythia-6.9b-deduped\_1shot}  \\

			\textbf{pythia-6.9b\_1shot}
        \end{tcolorbox}
        
        \begin{tcolorbox}[colback=agentpink!5!white,colframe=agentpink!75!black,title=Inst-tuned LLMs, fonttitle=\bfseries\tiny\ttfamily, coltitle=white, fontupper=\tiny\ttfamily]
			\textbf{flan-t5-xxl\_1shot}  \\

			\textbf{flan-ul2\_1shot}  \\

			\textbf{gemma-1.1-7b-it\_1shot}  \\

			\textbf{mistral-7b-inst\_1shot}  \\

			\textbf{opt-iml-max-30b\_1shot}  \\

			\textbf{phi-3-mini-3.8b\_1shot}
        \end{tcolorbox}
        
    \caption{Agents we use in the LLMs category.}\label{fig:appendix-agent-names-llms}
\end{figure}

\paragraph{Large Language Models (\llm{s}).}
We evaluate an array of \llm{s}, grouped below by their training / scale.
All models are evaluated in a zero-shot manner (no finetuning over \qb{} questions).

\noindent\textit{Base Models:} 
The models are exclusively trained on an unsupervised CausalLM objective: OPT~\cite{Zhang2022OPTOP}, GPT-Neo~\cite{Black2021GPTNeoLS} and Pythia~\cite{biderman2023pythia}

\noindent\textit{Benchmark Instruction Tuned (IT) Models:} \llm{s} fine-tuned on tasks
with natural instructions over each benchmark; OPT-IML~\cite{iyer2022optiml}, 
T0, T0pp~\cite{sanh2021multitask}, Flan-T5~\cite{Chung2022ScalingIL} and Flan-UL2~\cite{tay2022ul2}.

\noindent\textit{Very Large-Scaled Models:}
Llama-2 (70 billion parameters)~\cite{touvron2023llama}
and Falcon (40 billion parameters)~\cite{falcon40b} and its instruction tuned variant.
Due to limited information on their training data mixtures, direct comparisons with
other models are challenging. Nevertheless, we include these large-scale models to
gauge their performance relative to humans.

\noindent\textit{Closed-Sourced Model-Based APIs:} OpenAI's ChatGPT~\cite{NEURIPS2022_b1efde53}
and GPT-4 Turbo~\cite{openai2023gpt4}
\begin{figure}[h]
        \centering
    
        \begin{tcolorbox}[colback=agentpurple!5!white,colframe=agentpurple!75!black,title=OpenAI GPT3+, fonttitle=\bfseries\tiny\ttfamily, coltitle=white, fontupper=\tiny\ttfamily]
			\textbf{openai-gpt-3.5-turbo\_1shot}  \\

			\textbf{openai-gpt-4-turbo\_1shot}  \\

			\textbf{openai-gpt-4o\_1shot}
        \end{tcolorbox}
        
    \caption{Agents we use in the GPT-3+ category.}\label{fig:appendix-agent-names-gpt-3+}
\end{figure}

None of the Transformer-based models, including those pretrained on \qa{} datasets like
TriviaQA, are specifically finetuned on \qb{}; we adhere to the standard in-context 
learning practice~\cite{Brown2020LanguageMA},providing a task instruction followed
by concatenated \qa{} pair demonstrations.
\autoref{fig:appendix-prompt-llm} shows an example of the prompt used for these models.

\begin{figure}[h]
        \centering
    
        \begin{tcolorbox}[colback=agentdarkblue!5!white,colframe=agentdarkblue!75!black,title=RAG (Top 10), fonttitle=\bfseries\tiny\ttfamily, coltitle=white, fontupper=\tiny\ttfamily]
			\textbf{rag-bm25\_top10-flan-ul2}  \\

			\textbf{rag-bm25\_wiki\_top10-command-r-plus}  \\

			\textbf{rag-grit\_top10-flan-ul2}  \\

			\textbf{rag-grit\_wiki\_top10-command-r-plus}
        \end{tcolorbox}
        
        \begin{tcolorbox}[colback=agentdarkblue!5!white,colframe=agentdarkblue!75!black,title=RAG-flan-t5-xl (Top 3), fonttitle=\bfseries\tiny\ttfamily, coltitle=white, fontupper=\tiny\ttfamily]
			\textbf{rag-bm25\_top3-T0pp-11b}  \\

			\textbf{rag-bm25\_top3-flan-t5-xl}  \\

			\textbf{rag-contriever\_top3-T0pp-11b}  \\

			\textbf{rag-contriever\_top3-flan-t5-xl}
        \end{tcolorbox}
        
    \caption{Agents we use in the RAG category.}\label{fig:appendix-agent-names-rag}
\end{figure}

\paragraph{Retriever-augmented Generative Models.} Following the RAG paradigm from~\citep{lewis2020retrieval} 
for open-domain \qa{}, we first retrieve Wikipedia documents relevant to 
the questions, then employ a generator model for short answer generation.
Our retrievers include dense \abr{contriever} and a sparse passage retriever (BM25).
For the retriever, we use both a dense retriever (\abr{contriever}) as well as a
sparse passage retriever that uses BM25 to encode documents.
In our study, we mainly use FlanT5-XL~\cite{Chung2022ScalingIL} as the generator model,
whose input context is limited to 512 tokens and composed of the top-3 documents by retriever. 
We also explore Flan-UL2~\cite{tay2022ul2}, 
an instruction-tuned UL2 with a 2048-token receptive field, to handle all the 10 documents.
\autoref{fig:appendix-prompt-rag} shows an example of the prompt used for \abr{rag} models.

\definecolor{britishracinggreen}{rgb}{0.0, 0.26, 0.15}

\newcommand{\cluesuffix}{\textcolor{britishracinggreen}{What is being talked about here? Answer the question in a single word / short phrase.}}

\newcommand{\ragcluesuffix}{\textcolor{britishracinggreen}{What is being talked about here? Find the answer from above documents and answer in a single word or a short phrase.}}

\begin{figure}[ht]
  \centering
  \begin{framed}
  \begin{minipage}{\linewidth}
    \tiny{
        \textcolor{britishracinggreen}{\textvtt{You are a Quizbowl agent expert in Question Answering. Questions are in form of single or multiple clue(s) about a certain concept / entity. The following is a list of Quizbowl clues. Deduce the answer based on what the clues are describing, and answer the question in the form of a single word or a short phrase.}} \\
    
        \textvtt{\textcolor{britishracinggreen}{Question:} \textcolor{gray}{\{ demonstration clues \} \cluesuffix{}}}

        \textvtt{\textcolor{britishracinggreen}{Answer:} \textcolor{gray}{\{ demonstration answer \}}} \\

        \textvtt{\textcolor{britishracinggreen}{Question:} \textcolor{gray}{\{ inference clues \} \cluesuffix{}}}
        
        \textvtt{\textcolor{britishracinggreen}{Answer:}}
    }    
  \end{minipage}
  \end{framed}
\caption{A condensed version of our prompt to Base models, Instruction-tuned models and Closed-source models~(\S~\ref{subsec:ai-agents}).}
\label{fig:appendix-prompt-llm}
\end{figure}

\begin{figure}[ht]
  \centering
  \begin{framed}
  \begin{minipage}{\linewidth}
    \tiny{
        \textcolor{britishracinggreen}{\textvtt{You are a Quizbowl agent expert in Question Answering. Questions are in form of single or multiple clue(s) about a certain concept / entity. Answer the Quizbowl question by finding a short answer from the reference documents listed below.}} \\
    
        \textvtt{\textcolor{britishracinggreen}{Documents:}}

        \textvtt{\textcolor{gray}{\{ Document 1 Title\}}:}
        \textvtt{\textcolor{gray}{\{ Document 1 Content\}}}

        \textvtt{\textcolor{gray}{\{ Document 2 Title\}}:}
        \textvtt{\textcolor{gray}{\{ Document 2 Content\}}}

        \textvtt{\textcolor{gray}{$\ldots$}}

        \textvtt{\textcolor{gray}{\{ Document k Title\}}:}
        \textvtt{\textcolor{gray}{\{ Document k Content\}}} \\

        \textvtt{\textcolor{britishracinggreen}{Question:} \textcolor{gray}{\{ inference clues \} \ragcluesuffix{}}}
        
        \textvtt{\textcolor{britishracinggreen}{Answer:}}
    }    
  \end{minipage}
  \end{framed}
\caption{A condensed version of our prompt to our retriever-augmented generative (\abr{rag}) models~(\S~\ref{subsec:ai-agents}).}
\label{fig:appendix-prompt-rag}
\end{figure}

\paragraph{Answer Match Evaluation.}
Traditional exact-match metric often misses alternative answers that have different wordings or forms but the same semantic meaning as the correct answer~\cite{bulian2022tomayto}. 
%
%
To better handle this, we adopt a fuzzy match evaluation using multiple-answer
aliases~\cite{Si2021WhatsIA}: if the character level 
matching rate between the predicted answer and the gold answer exceeds a certain threshold, the prediction is considered as correct.
%
The threshold is tuned against human judgments on a small development set.
\section{Question Features for Logistic Regression Study}\label{appendix:logreg-feats}
This section describes the features used in the logistic regression study in \S~\ref{subsec:kira-setup}.

\paragraph{Question Category Features.} These features are binary and indicate whether a
question belongs to a specific category. These categories are the one highlighted in~\autoref{fig:question-categories}. 
The categories are: {\tiny\texttt{c\_question\_categories}, \texttt{c\_fine\_arts}, \texttt{c\_cultural\_geography}, \texttt{c\_geography}, \texttt{c\_physical\_geography}, \texttt{c\_political\_geography}, \texttt{c\_technical\_geography}, \texttt{c\_ancient\_history}, \texttt{c\_history}, \texttt{c\_cultural\_history}, \texttt{c\_exploration\_and\_colonization}, \texttt{c\_military\_history}, \texttt{c\_other}, \texttt{c\_political\_history}, \texttt{c\_scientific\_history}, \texttt{c\_social\_history}, \texttt{c\_language}, \texttt{c\_author\_and\_works}, \texttt{c\_literature}, \texttt{c\_genre\_and\_style}, \texttt{c\_literary\_terms}, \texttt{c\_plot\_and\_characters}, \texttt{c\_music}, \texttt{c\_mythology}, \texttt{c\_political\_events}, \texttt{c\_politics}, \texttt{c\_political\_figures}, \texttt{c\_political\_institutions}, \texttt{c\_political\_theory}, \texttt{c\_religion}, \texttt{c\_astronomy}, \texttt{c\_science}, \texttt{c\_biology}, \texttt{c\_chemistry}, \texttt{c\_earth\_science}, \texttt{c\_materials}, \texttt{c\_mathematics}, \texttt{c\_other}, \texttt{c\_physics}, \texttt{c\_scientific\_history}, \texttt{c\_sports}, \texttt{c\_technology}, \texttt{c\_television/movies}}

\paragraph{Linguistic Features}
\textit{LingFeat} is a Python research package designed for the extraction of various handcrafted linguistic features, positioning itself as a comprehensive \nlp{} feature extraction tool. Currently, it is capable of extracting 255 linguistic features from English textual inputs.
The features extracted by \textit{LingFeat} span across five broad linguistic branches that \citet{lee-etal-2021-pushing} details.
\begin{itemize}
    \item \textbf{Advanced Semantic (AdSem):} Aims at measuring the complexity of meaning structures. Note: This feature is currently facing some operational issues, which are under investigation.
    \item \textbf{Semantic Richness, Noise, and Clarity:} Extracted from trained LDA models. The models are included and require no further training.
    \item \textbf{Discourse (Disco):} Focuses on measuring coherence and cohesion through entity counts, entity grid, and local coherence score.
    \item \textbf{Syntactic (Synta):} Evaluates the complexity of grammar and structure, including phrasal counts (e.g., Noun Phrase), part-of-speech counts, and tree structure.
    \item \textbf{Lexico Semantic (LxSem):} Measures word/phrasal-specific difficulty through metrics like type-token ratio, variation score (e.g., verb variation), age-of-acquisition, and SubtlexUS frequency.
    \item \textbf{Shallow Traditional (ShTra):} Encompasses traditional features/formulas for assessing text difficulty, such as basic average counts (words per sentence), Flesch-Kincaid Reading Ease, Smog, Gunning Fog, etc.
\end{itemize}

\paragraph{Time based features}
We create two time based feature, {\small\texttt{t\_range}} and {\small\texttt{t\_range}}. Both are binary features. {\small\texttt{t\_range}} is 1 if the question was asked in the context of certain time period or a range, (e.g., \textit{in the 20th century}, \textit{in the 19th}), and 0 otherwise. {\small\texttt{t\_range}} is 1 if the question refers to an event related to another event, (e.g., \textit{after the fall of Rome}, \textit{before the French Revolution}), and 0 otherwise.

\paragraph{Other features}
{\small\texttt{o\_TRASH}} is 1 is the question enquires about specific events in pop culture category, and 0 otherwise. This feature reflects the TRASH category from Quizbowl. Similarly,
{\small\texttt{o\_Records}} is 1 if the question enquires about specific records through mention of superlative forms of words like ``most recent'', ``best category'', etc, and 0 otherwise. This feature reflects the Records category from Quizbowl.
\clearpage
\section{Question Difficulty}\label{appendix:question-difficulty}

This section enlists the full set of heatmaps of mean relevance $\rel{j,k}$ and \emph{mean effective difficulty} $\ediff{D, \mu_k}$
of question clusters across the five latent factors ($k$).

\begin{figure*}[th]
    \centering
    \includegraphics[width=\linewidth]{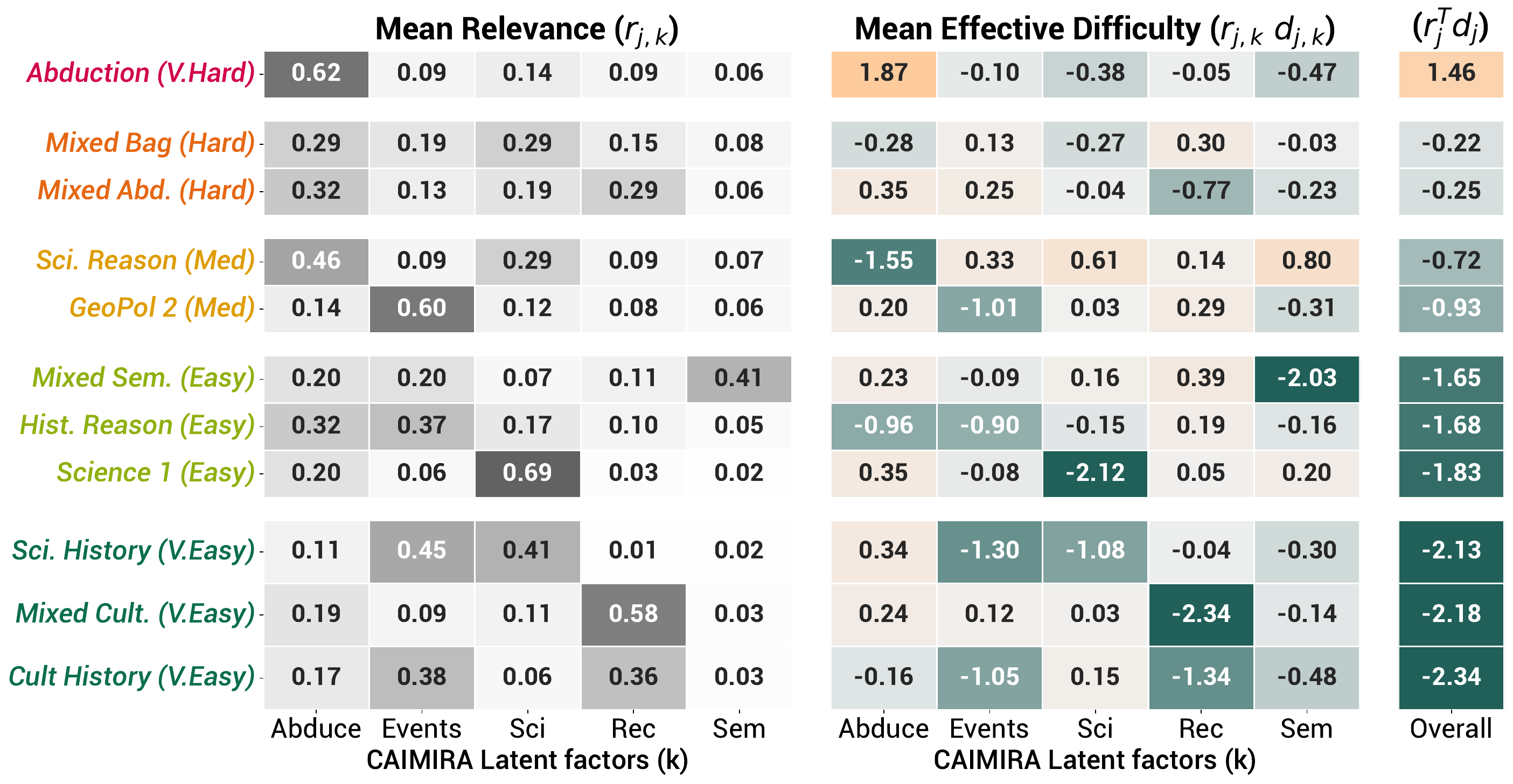}
    \caption{Heatmaps of mean relevance $\rel{j,k}$ and \emph{mean effective difficulty} $\ediff{D, \mu_k}$
    of question clusters across the five latent factors ($k$).
    }\label{fig:effective-difficulty-appendix}
  \end{figure*}

  \begin{figure*}[th]
    \centering
    \includegraphics[width=\linewidth]{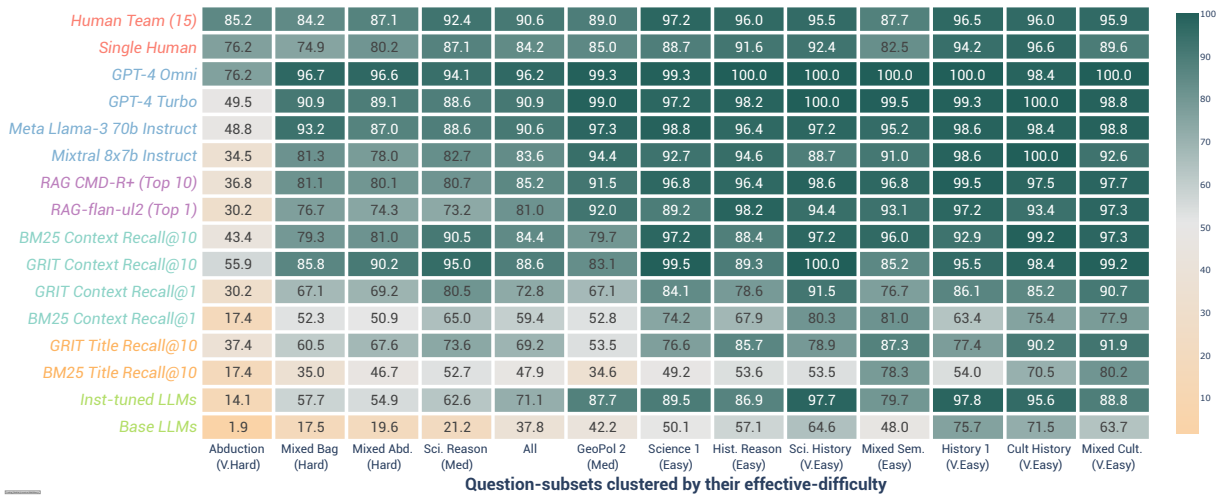}
    \caption{Full set of agent accuracies across all question clusters defined in \autoref{fig:effective-difficulty-appendix}.
    We use the same color scheme as in \autoref{fig:agent-accuracies}.}\label{fig:appendix-agent-accuracies}
  \end{figure*}
\definecolor{vhard}{HTML}{d1064a}
\definecolor{hard}{HTML}{e66712}
\definecolor{med}{HTML}{de9d04}
\definecolor{easy}{HTML}{90b010}
\definecolor{veasy}{HTML}{0B6E4F}

\begin{figure*}[ht]
        \centering
    
        \begin{tcolorbox}[colback=vhard!5!white,colframe=vhard!75!black,title=Abduction (V.Hard), fonttitle=\bfseries\small\ttfamily, coltitle=white, fontupper=\tiny\ttfamily]
    		\underline{\textbf{Answer: Mount Olympus}} \\ Clues: Homer claimed that this place never has storms and is bound in aether. \\

		\underline{\textbf{Answer: medians}} \\ Clues: Apollonius' Theorem can be used to find the length of this construct given the side lengths of a triangle. \\

		\underline{\textbf{Answer: {The Arnolfini Marriage}}} \\ Clues: Symbols in this painting include a pair of discarded clogs and a chandelier with one lit candle. In the middle of this painting, a feather duster and a beaded chain flank the artist's signature, which is above a circular mirror. A dog sits near this painting's two human figures, one of whom wears a green dress as she holds the hand of her suitor.(*) \\

		\underline{\textbf{Answer: {Ramona} Geraldine {Quimby}}} \\ Clues: This owner of a stuffed elephant named Ella Funt plays a black-nosed sheep in a Christmas play and dresses up as "the baddest witch in the world." She has a cat named Picky-Picky until it dies, and she also sees herself in an infinite mirror. \\

		\underline{\textbf{Answer: A Wrinkle in Time}} \\ Clues: Two characters in this book later appear as the main characters of Many Waters. Mrs. Whatsit, Mrs. Who, and Mrs.Which start this journey in this book. \\

		\underline{\textbf{Answer: {rectangles}}} \\ Clues: The uniform probability distribution takes this shape. Rotating this shape using one of its sides as an axis yields a cylinder. This shape is traced out by the x-axis, the y-axis, and the equations x equals two and y equals six. \\

		\underline{\textbf{Answer: To Kill a Mockingbird}} \\ Clues: One character in this book deliberately pours syrup all over his lunch. At one point, the main characters are taken to a church by their cook, Calpurnia. \\

		\underline{\textbf{Answer: (Alexandre) Gustave {Eiffel}}} \\ Clues: This man designed railway stations in Santiago, Chile and Budapest, Hungary. He was jailed after being implicated in a failed Panama Canal project, for which he designed the locks. \\

		\underline{\textbf{Answer: Lord of the Flies}} \\ Clues: In this novel, a dead parachutist is discovered by the strange introverted character Simon. Sam and Eric are the last followers of one character in this novel. \\

		\underline{\textbf{Answer: Eminem}} \\ Clues: This musician says, after declaring "now I'm gonna make you dance," "girl you know you're my world" in his song "Just Lose It." 
        \end{tcolorbox}
        
        \caption{Examples of questions from different clusters.}
        \label{fig:appendix-question-examples-cluster-0}
    \end{figure*}

\begin{figure*}[ht]
        \centering
    
        \begin{tcolorbox}[colback=vhard!5!white,colframe=vhard!75!black,title=Mixed Abd. (Hard), fonttitle=\bfseries\small\ttfamily, coltitle=white, fontupper=\tiny\ttfamily]
    		\underline{\textbf{Answer: Justin Bieber}} \\ Clues: This singer claims "I'd wait for you forever and a day" and "your world is my world" in one song. Big Sean wonders "I don't know if this makes sense, but you're my hallelujah" in a song where this singer says he'll be your (*) platinum, silver and gold. \\

		\underline{\textbf{Answer: Neil {Gaiman}}} \\ Clues: This frequent collaborator of Dave McKean won both the Carnegie and Newbery Medals for a book about a crypt full of Sleer being explored by Nobody Owens. \\

		\underline{\textbf{Answer: {Moby}-{Dick} (or {The Whale})}} \\ Clues: Characters in this novel include the Zoroastrian Fedallah (feh-DAH-lah), a Native American called Tashtego, and a South Sea islander named Queequeg (KWEE-KWAIG). \\

		\underline{\textbf{Answer: {Samson}}} \\ Clues: Before he was born, his parents learned that he was not to touch a dead body, and he was to abstain from strong drink. He was involved with a Timnite woman and a harlot before meeting the woman that would betray him. \\

		\underline{\textbf{Answer: Aeneas}} \\ Clues: This man is told by the ghost of his wife Creusa to leave for Hesperia after carrying his father Anchises (ann-KYE-sees) and son Ascanius out of a besieged city. He visits the underworld with the help of a golden bough, on the advice of the Cumaean Sibyl. \\

		\underline{\textbf{Answer: {Mean}}} \\ Clues: The harmonic one of n numbers in a data set is n divided by the sum of the reciprocals of the numbers. The geometric one is the nth root of the product of the numbers. The geometric one is always less than or equal to the arithmetic ("air-ith-MET-ick") one. \\

		\underline{\textbf{Answer: {Alice}}} \\ Clues: This character watches a lion and a unicorn fight over a crown, and although her cat Dinah will not talk to her, the Tiger Lily and the other flowers will. \\

		\underline{\textbf{Answer: {Daniel}}} \\ Clues: As punishment for not worshipping a golden statue, this man's friends were ordered thrown into a furnace, but they were not burned. While training to be a scribe, this man was given the Babylonian name Belteshazzar (“BEL-tuh-SHAH-zar”). \\

		\underline{\textbf{Answer: {magma}}} \\ Clues: The three types of this material differ by their mineral and gas content; rhyolitic and andesitic types contain more silicon dioxide and are more viscous. The basaltic type is hottest, forms due to partial melting in the mantle, and flows fastest. \\

		\underline{\textbf{Answer: {parallelogram}}} \\ Clues: This shape names a law for adding vectors. In a namesake illusion, diagonals of two of these figures appear to be different lengths, though they are not. 
        \end{tcolorbox}
        
        \caption{Examples of questions from different clusters.}
        \label{fig:appendix-question-examples-cluster-1}
    \end{figure*}

\begin{figure*}[ht]
        \centering
    
        \begin{tcolorbox}[colback=hard!5!white,colframe=hard!75!black,title=Mixed Bag (Hard), fonttitle=\bfseries\small\ttfamily, coltitle=black, fontupper=\tiny\ttfamily]
    		\underline{\textbf{Answer: {prime} numbers}} \\ Clues: The fundamental theorem of arithmetic states that every positive integer can be uniquely represented as a product of these numbers. Special types of these numbers are named after Fermat (“fur-MAHT”) and Mersenne (“mur-SEN”). To find these numbers, one may use the Sieve of Eratosthenes (air-uh-TOSS- then-eez”), in which one crosses off all multiples of two, then all multiples of three, and so on. For 10 points, give these numbers whose only factors are one and themselves.\\

		\underline{\textbf{Answer: gerrymandering}} \\ Clues: The Justice Department suggested using race as a basis for this practice in the 1990's. \\

		\underline{\textbf{Answer: Secretary of State}} \\ Clues: Resignations of the President or Vice-President must be delivered to this person. Madeleine Albright was the first woman to hold this position, and one candidate for this position in the second Obama administration withdrew her candidacy due to controversy over the (*) Benghazi attacks. \\

		\underline{\textbf{Answer: {Romeo} and {Juliet}}} \\ Clues: This play's opening brawl is started by Gregory and Samson. Later in this play, Friar John fails to deliver a letter written by Friar Lawrence. \\

		\underline{\textbf{Answer: Sagittarius}} \\ Clues: Both Globular Cluster M54, the center of this constellation's namesake dwarf elliptical galaxy, and a possible supermassive black hole at the center of the Milky Way are found in this constellation. \\

		\underline{\textbf{Answer: photographs}} \\ Clues: An early invention used to make art works in this medium was the daguerreotype [duh-gayr-"row"-"type"]. Eadweard ["edward"] Muybridge created works in this medium which clarified the method by which horses gallop. The Steerage and Migrant Mother are specific examples of these types of art works. \\

		\underline{\textbf{Answer: {sine}}} \\ Clues: This function's namesake law relates the side length to the opposite angle in any triangle. \\

		\underline{\textbf{Answer: static}} \\ Clues: This term describes a type of friction whose coefficient is usually larger than that of kinetic friction. It describes a type of equilibrium in which the net torque and net force both equal zero, resulting in a motionless object. \\

		\underline{\textbf{Answer: {Peter I}}} \\ Clues: This man's reign began with the Streltsy (SHTRELT-zee) Revolt instigated by his half-sister, Sophia. \\

		\underline{\textbf{Answer: {greatest common factor}}} \\ Clues: Antenaresis, or Euclid's method, can be used to find this value given any two numbers. It can be also be found by multiplying two numbers and dividing by their least common multiple. 
        \end{tcolorbox}
        
        \caption{Examples of questions from different clusters.}
        \label{fig:appendix-question-examples-cluster-2}
    \end{figure*}

\begin{figure*}[ht]
        \centering
    
        \begin{tcolorbox}[colback=hard!5!white,colframe=hard!75!black,title=Sci. Reason (Med), fonttitle=\bfseries\small\ttfamily, coltitle=black, fontupper=\tiny\ttfamily]
    		\underline{\textbf{Answer: 2}} \\ Clues: Euler characteristic of platonic solids have this value. This integer times pi gives the number of radians in the unit circle. Truth tables can evaluate to this many outputs. \\

		\underline{\textbf{Answer: tundra}} \\ Clues: Cushion plants are found in the alpine form of this biome, which is also home to marmots, pikas, and chinchillas. The point at which this biome meets taiga is known as the treeline. Flora in this biome consists of lichens (LYE-kens) and mosses. Non-alpine forms of it have little vegetation due to permafrost. \\

		\underline{\textbf{Answer: Lois {Lowry}}} \\ Clues: One of this writer's stories follows Annemarie Johansen as she helps her friend Ellen escape from Nazi-occupied Denmark. A sequel to this author's most well-known book follows the weaver Kira, and that book ends with Jonah and Gabe fleeing the dystopian society they live in. \\

		\underline{\textbf{Answer: calcium}} \\ Clues: Channels that carry ions made of this element are blocked by some hypertension medications. \\

		\underline{\textbf{Answer: \"My Life Would Suck Without You\"}} \\ Clues: The protagonists of this song's music video throw magazines, clothes and an empty fishbowl out an open window. This song notes that "maybe I was stupid for telling you goodbye" regarding a boy who the singer supposes is sorry because "you're (*) standing at my door." This song's chorus notes that "you've got a piece of me and honestly" before expressing the title sentiment. \\

		\underline{\textbf{Answer: {Ramona} Geraldine {Quimby}}} \\ Clues: This owner of a stuffed elephant named Ella Funt plays a black-nosed sheep in a Christmas play and dresses up as "the baddest witch in the world." She has a cat named Picky-Picky until it dies, and she also sees herself in an infinite mirror. This best friend of Howie Kemp lives on the same street as Henry Higgins. For 10 points, name this little sister of Beezus, the main character of a series of books by Beverly Cleary.\\

		\underline{\textbf{Answer: guns}} \\ Clues: In Major Barbara, Andrew Undershaft became rich by manufacturing these objects. Both Hedda Gabler and Young Werther (VEHR-tuhr) commit suicide using these objects. \\

		\underline{\textbf{Answer: Bridge to Terabithia}} \\ Clues: This novel's protagonist wants to become the fastest runner in the fifth grade, but that plan is spoiled by the girl who moves in next door. While this book's protagonist visits the National Art Gallery with his music teacher, that girl tries to (*) swing over the creek, but the rope snaps and she dies. \\

		\underline{\textbf{Answer: Curie}} \\ Clues: Two brothers of this surname discovered piezoelectricity and a namesake point at which ferromagnetic materials become paramagnetic. One of those brothers explored the properties of the ore pitchblende with his wife. That wife later won a second Nobel Prize for her work isolating radium, and named the element polonium after her native country. For 10 points, give the last name of physicist Pierre and his wife Marie.\\

		\underline{\textbf{Answer: {polls}}} \\ Clues: The “straw” form of this practice is unscientific and the “push” form of this is really just a campaign tactic designed to attack an opponent in disguise. 
        \end{tcolorbox}
        
        \caption{Examples of questions from different clusters.}
        \label{fig:appendix-question-examples-cluster-3}
    \end{figure*}

\begin{figure*}[ht]
        \centering
    
        \begin{tcolorbox}[colback=easy!5!white,colframe=easy!75!black,title=Mixed Sem. (Easy), fonttitle=\bfseries\small\ttfamily, coltitle=black, fontupper=\tiny\ttfamily]
    		\underline{\textbf{Answer: Richard I of England}} \\ Clues: This man was killed by a crossbow bolt while besieging the castle Charlus-Chabrol. After the departure of Philip Augustus of France, this man led the Christian armies in the Third Crusade, during which he achieved peace with Saladin. He was succeeded by his brother John. For 10 points, name this 12th-century King of England known by an epithet signifying his bravery.\\

		\underline{\textbf{Answer: Vincent (Willem) {Van Gogh}}} \\ Clues: While in Auvers [oh-vair], this man painted his physician holding a foxglove plant. In another painting by him, a woman pours coffee as a destitute family sits at a table for a meal. His best-known work shows Saint- Rémy [sahn-ray-mee], and this artist painted the Portrait of Dr. Gachet [gah-shay] and The Potato Eaters. \\

		\underline{\textbf{Answer: William {Faulkner}}} \\ Clues: In this author's first Pulitzer Prize-winning work, the Generalissimo orders the execution of Corporal Zsettslani (“SET-slah-nee”). His second Pulitzer-winning novel revolves around Lucius Priest, a resident of Yoknapatawpha (“YOCK-NAH-puh-TAH-fuh”) County. This author wrote novels about Thomas Sutpen and about the death of Addie Bundren. For 10 points, name this American author of Absalom! Absalom!, As I Lay Dying, and The Sound and the Fury.\\

		\underline{\textbf{Answer: Antonio López de {Santa Anna}}} \\ Clues: This figure ordered the Goliad Massacre, and he was severely injured by French cannon fire at Veracruz during the Pastry War. The Treaties of Velasco were signed following this leader's capture after the Battle of San Jacinto, and he was responsible for the deaths of Jim Bowie and Davy Crockett. \\

		\underline{\textbf{Answer: "Auld Lang Syne"}} \\ Clues: This poem's original form notes that the speaker and his addressee have "rin about the braes" and "paidl't i' the burn." The speaker of this poem written in Scottish dialect claims that they will "take a cup of kindness yet" and asks, "Should auld acquaintance be forgot, and never brought to min'?" For 10 points, name this Robert Burns poem that is often sung on New Year's Eve.\\

		\underline{\textbf{Answer: Pytor Ilyich {Tchaikovsky}}} \\ Clues: This musician dedicated his Symphony No. 4 in F Minor to his financial supporter Nadezhda (nah- DEZH-dah) von Meck, though they never met. His Sixth Symphony, nicknamed Pathetique (pah-theh- TEEK), premiered nine days before his death. \\

		\underline{\textbf{Answer: The Outsiders}} \\ Clues: In this novel, Bob Sheldon and Randy Adderson take part in an attack on Johnny, causing Johnny to fear for his life. \\

		\underline{\textbf{Answer: To Kill a Mockingbird}} \\ Clues: In this novel the narrator's father shoots Tim Johnson, a rabid dog. The narrator and her brother are attacked on the way home from a Halloween pageant, but are saved by Boo Radley. \\

		\underline{\textbf{Answer: {Johann Sebastian Bach}}} \\ Clues: Lieschen [lee-shen] is addicted to coffee in a cantata by this composer of the Notebook for Anna Magdalena. Gounod's [goo-noh's] Ave Maria is based on a prelude from this composer's Well-Tempered Clavier, and Mendelssohn revived his setting of the St. Matthew Passion. \\

		\underline{\textbf{Answer: {Don Quixote} de la Mancha}} \\ Clues: This character interrupts a round of storytelling by attacking a stash of wine-skins. He wears a washbasin as a helmet while calling himself the Knight of the Sorry Face. He owns the horse Rocinante (ROHsin- AHN-tay) and frequently speaks of his love for Dulcinea (dull-sin-AY-ah) to his friend Sancho Panza. For 10 points, name this self-proclaimed knight from La Mancha who fights against windmills in a book by Miguel de Cervantes.
        \end{tcolorbox}
        
        \caption{Examples of questions from different clusters.}
        \label{fig:appendix-question-examples-cluster-5}
    \end{figure*}

\begin{figure*}[ht]
        \centering
    
        \begin{tcolorbox}[colback=easy!5!white,colframe=easy!75!black,title=Science 1 (Easy), fonttitle=\bfseries\small\ttfamily, coltitle=black, fontupper=\tiny\ttfamily]
    		\underline{\textbf{Answer: {Spanish}}} \\ Clues: One writer in this language wrote the collection “Twenty Love Poems and a Song of Despair.” \\

		\underline{\textbf{Answer: {Earth}}} \\ Clues: In Jainism, this object's central point is Mount Meru. In Chinese mythology, this object is the lower half of a cosmic egg split by Pangu, while in ancient Egypt the original form of this object was the primordial (*) mound. \\

		\underline{\textbf{Answer: {mitochondria} (“ MY-toe-KON-dree-uh ”)}} \\ Clues: The DNA in this organelle (“or-guh-NELL”) is inherited only from the mother. The inner membrane of this organelle contains folds known as cristae (“CRISS-tay”) and encloses its matrix. \\

		\underline{\textbf{Answer: {coral reefs}}} \\ Clues: Darwin's first paper was on the formation of this biome, whose organisms are threatened by white-band disease. Acidification removes the minerals needed for this ecosystem to grow as each new generation builds on the calcium carbonate skeletons of the previous one. \\

		\underline{\textbf{Answer: Ohio}} \\ Clues: n this state's capital, the Lane Avenue Bridge crosses the Olentangy River. Another of its cities contains historic Italian architecture in its Over-the-Rhine neighborhood, while another city, at the mouth of the Cuyahoga River, contains Case Western Reserve University. Much of its northern border is at Lake (*) Erie, and it is separated from Kentucky by its namesake river. For 10 points, name this state containing Cincinnati, Cleveland, and Columbus.\\

		\underline{\textbf{Answer: {Chlorine} or {Cl}}} \\ Clues: Stomach acid consists mainly of a compound of hydrogen and this element. It is the second-lightest halogen, after fluorine, and at room temperature is a yellow-green gas. Compounds with it, carbon, hydrogen, and fluorine deplete the ozone layer and are called (*) CFCs. It is used in bleach as well as to disinfect swimming pools, and forms table salt along with sodium. For 10 points, name this element, number 17, symbolized Cl.\\

		\underline{\textbf{Answer: electron}} \\ Clues: This particle was discovered by J.J. Thomson, and its exact charge was discovered in the Millikan oil drop experiment. According to the Pauli Exclusion Principle, two of these particles cannot exist in the same quantum state. \\

		\underline{\textbf{Answer: matter}} \\ Clues: The density parameter for the non-relativistic form of this falls off with the cube of the scale factor. This substance dominated the universe from approximately 75,000 years after the Big-Bang until about 4 billion years ago. \\

		\underline{\textbf{Answer: {violin}}} \\ Clues: The Rhapsody on a Theme of Paganini was written from twenty-four caprices originally written for this instrument. Vivaldi's The Four Seasons is a set of concerti (“con-CHAIR-tee”) written for this instrument. \\

		\underline{\textbf{Answer: glaciers}} \\ Clues: These objects contain the zone of plastic flow and the zone of brittle flow. They are formed by compressing firn, and parts of them break off by calving. Till is soil left behind by these objects, which also push material to form moraines. 
        \end{tcolorbox}
        
        \caption{Examples of questions from different clusters.}
        \label{fig:appendix-question-examples-cluster-6}
    \end{figure*}

\begin{figure*}[ht]
        \centering
    
        \begin{tcolorbox}[colback=easy!5!white,colframe=easy!75!black,title=Hist. Reason (Easy), fonttitle=\bfseries\small\ttfamily, coltitle=black, fontupper=\tiny\ttfamily]
    		\underline{\textbf{Answer: Scooby-Doo}} \\ Clues: Big Bob Oakley was the first person on this show to say "I'd have gotten away with it too, if it weren't for those kids," and one show in this series introduced a character named Scrappy. In 2002, a film of the same name starred Freddie Prinze, Jr. as Freddy and Sarah Michelle Gellar as Daphne. For 10 points, name this cartoon franchise, named for a cowardly Great Dane.\\

		\underline{\textbf{Answer: Steve Jobs}} \\ Clues: This man, along with Edwin Catmull, was credited as an executive producer of the original Toy Story movie, produced by Pixar Animation, which he renamed after purchasing it from George Lucas in 1986. From 2000 to 2011, he served as CEO of the computer company he co-founded with Steve Wozniak. \\

		\underline{\textbf{Answer: Neptune}} \\ Clues: A triangular patch of clouds that circulates this planet quickly is known as The Scooter. Its atmosphere contains the fastest winds in the solar system. Its existence was predicted by Alexis Bouvard, and it was discovered by Johann Galle. It often contains the Great Dark Spot. Its largest moon, which has a retrograde orbit, is Triton. For 10 points, name this gas giant, the farthest from the Sun in the solar system.\\

		\underline{\textbf{Answer: Orion}} \\ Clues: This constellation contains the Trapezium Cluster and is the site of a late-October meteor shower. \\

		\underline{\textbf{Answer: Niccolo {Machiavelli}}} \\ Clues: Although he is not Sun Tzu, this man wrote a version of The Art of War. He wrote a critique of Roman history in his Discourses on Livy. \\

		\underline{\textbf{Answer: {prime} numbers}} \\ Clues: The fundamental theorem of arithmetic states that every positive integer can be uniquely represented as a product of these numbers. \\

		\underline{\textbf{Answer: The {New York Times}}} \\ Clues: This newspaper was sued by Alabama public safety officer Louis B. Sullivan. Its long-time publisher, Arthur Ochs Sulzberger, died in 2012. \\

		\underline{\textbf{Answer: Uncle Tom's Cabin}} \\ Clues: In this novel, shelter is provided by the Halliday and Bird families. At the beginning of this novel, the Shelby family sells their property to the St. Clare family. At the end of this novel, George and Eliza Harris escape north. The husband of Aunt Chloe is killed by Simon Legree in, for 10 points, what American novel, depicting the life of slaves, written by Harriet Beecher Stowe?\\

		\underline{\textbf{Answer: Harry Mason {Reid}}} \\ Clues: This man almost lost his Senate seat in the 1998, surviving a challenge from future colleague John Ensign, and he is expected to have a tough re-election in 2010 against Sue Lowden or Danny Tarkanian. He commented that Barack Obama was “light-skinned” and “spoke with no Negro dialect, unless he wanted one.” For 10 points, name this senior Senator from Nevada, the current Senate Majority Leader.\\

		\underline{\textbf{Answer: Pangaea}} \\ Clues: One piece of evidence that supports its existence is that the Caledonian mountains of Northern Europe are a continuation of the Appalachian Mountains. This entity broke up into Laurasia and Gondwanaland (“gon-DWON-uh-land”). 
        \end{tcolorbox}
        
        \caption{Examples of questions from different clusters.}
        \label{fig:appendix-question-examples-cluster-7}
    \end{figure*}

\begin{figure*}[ht]
        \centering
    
        \begin{tcolorbox}[colback=easy!5!white,colframe=easy!75!black,title=History 1 (V.Easy), fonttitle=\bfseries\small\ttfamily, coltitle=black, fontupper=\tiny\ttfamily]
    		\underline{\textbf{Answer: Puerto Rico}} \\ Clues: The independence of this commonwealth has been sought by Rubén Berríos, while an opposite approach has been pushed by its New Progressive Party under Pedro Pierluisi. In 2012, this commonwealth elected Alejandro García Padilla as governor and voted in a referendum to end its territorial status. (*) For 10 points, name this Caribbean Island, a United States territory that may someday become the 51st state.\\

		\underline{\textbf{Answer: {Philadelphia,} Pennsylvania}} \\ Clues: In this city, Wissahickon Creek goes through Fairmount Park. This city can be entered by crossing the Delaware River on the Betsy Ross Bridge. One of its buildings, where the Second Continental Congress adopted the (*) Declaration of Independence, is Independence Hall. The Liberty Bell is found in, for 10 points, what city in Pennsylvania?\\

		\underline{\textbf{Answer: Yellowstone National Park}} \\ Clues: The last wild herd of bison in the United States was located in this park, where today they are hunted by grizzly bears and wolves reintroduced in the 1990s. \\

		\underline{\textbf{Answer: Leo {Tolstoy}}} \\ Clues: One work by this author, about a man who injures himself while hanging curtains, is The Death of Ivan Ilyich. One of his novels has a relationship between Levin and Kitty, while the title character has an affair with Count Vronsky and eventually commits suicide by jumping in front of a (*) train. For 10 points, name this author who wrote about the French invasion of Russia in War and Peace in addition to writing Anna Karenina.\\

		\underline{\textbf{Answer: Federal Republic of {Germany}}} \\ Clues: One leader of this country forcibly annexed the Sudetenland (“soo-DAY-ten-land”). During a movement to reunite this country, the leader of one half operated under the policy of ostpolitik (“OST-pol- it-ick”). Following World War I, the Weimar (“VIE-mar”) Republic was established in this nation. \\

		\underline{\textbf{Answer: Thomas Jefferson}} \\ Clues: This politician responded to Francois Barbe-Marbois in his Notes on the State of Virginia. This man founded the University of Virginia and designed the mansion of Monticello.. \\

		\underline{\textbf{Answer: Mexico}} \\ Clues: In 1822, the House of Iturbide (“EE-tur-BEE-day”) assumed control of this nation for one year. This nation was ruled by an Austrian emperor installed by Napoleon III, Maximilian, although he was overthrown by Benito Juarez (“WAHR-ezz”). The Gadsden Purchase bought land from this country, whose victory at Puebla (“PWAY-bluh”) is celebrated as Cinco de Mayo. For 10 points, identify this nation that once owned California and Texas.\\

		\underline{\textbf{Answer: Ronald (Wilson) Reagan}} \\ Clues: This man used powers granted by the Taft-Hartley Act during a confrontation with air traffic controllers, and his Defense Secretary resigned after violations of the Boland Amendment were revealed. Before those events during his presidency, he served as Governor of California from 1967 until 1975. Prior to entering politics, this man was a famous (*) Hollywood actor. For 10 points, name this Republican president from 1981 to 1989.\\

		\underline{\textbf{Answer: Isaac Asimov}} \\ Clues: This author wrote a story in which the inhabitants of Lagash experience darkness for the first time. Along with "Nightfall," this author wrote a series of novels featuring the investigative interactions of Elijah Baley and R. Daneel Olivaw. Hari Selden invents the science of psychohistory in this author's novel (*) Foundation. For 10 points, name this Russian-American science fiction writer who depicted the Three Laws of Robotics in his collection, I, Robot.\\

		\underline{\textbf{Answer: Julius {Caesar}}} \\ Clues: This man fought against Ariovistus (“air-ee-oh-VIS-tuss”), a German leader, and Vercingetorix (“ver- KING-uh-TOR-ix”), a chieftain of the Arverni (“ar-VEHR-nee”) whose defeat is described in this man's book, Commentaries on the Gallic Wars. He led his troops across the Rubicon to start a civil war with Pompey, one of his partners in the First Triumvirate. For 10 points, name this Roman leader who was assassinated by Brutus on the Ides of March.
        \end{tcolorbox}
        
        \caption{Examples of questions from different clusters.}
        \label{fig:appendix-question-examples-cluster-8}
    \end{figure*}

\begin{figure*}[ht]
        \centering
    
        \begin{tcolorbox}[colback=veasy!5!white,colframe=veasy!75!black,title=Mixed Cult. (V.Easy), fonttitle=\bfseries\small\ttfamily, coltitle=white, fontupper=\tiny\ttfamily]
    		\underline{\textbf{Answer: The {Nutcracker}}} \\ Clues: This work opens with the title item given as a gift by Drosselmeyer; it is later broken by Fritz. Spanish, Arabian, and Chinese dances in this ballet are said to represent different substances such as chocolate, coffee, and tea. The Waltz of the Snowflakes and Dance of the (*) Sugarplum Fairy appear in, for 10 points, what Peter Tchaikovsky ballet about Clara's Christmas gift coming to life?\\

		\underline{\textbf{Answer: King {Arthur}}} \\ Clues: A popular novel about this figure is T.H. White's The Once and Future King. In the Annales Cambriae (ah-NAH-less CAM-bree-ay), this figure was mortally wounded at the Battle of Camlann during a fight with his son Mordred. \\

		\underline{\textbf{Answer: Thebes}} \\ Clues: This city was founded by Cadmus after following a cow until it sat. This city was besieged by the Sphinx, as all travelers who entered it were forced to either solve its riddle or be eaten. To avenge the sleight done to him by Eteocles(“et-TEE-oh-clees”), Polyneices (“polly-NYE-kees”) led a group of seven warriors against this city. \\

		\underline{\textbf{Answer: WikiLeaks}} \\ Clues: A PowerPoint presentation released by this organization details how Bank of America plans to attack it. One portion of this organization is run by the Sunshine Press. In November 2010, a Fox News host called it a "terrorist organization" after it published U.S. State Department diplomatic cables. \\

		\underline{\textbf{Answer: Isaac Newton}} \\ Clues: In this scientist's book Opticks, he discussed his experiments with the dispersion of light, including breaking white light into its constituent colors using a prism. One law named for him describes "universal (*) gravitation"; another states that the net force on an object is its mass times its acceleration, while a third states that for every action there is an equal and opposite reaction. For 10 points, name this English scientist who formulated three laws of motion.\\

		\underline{\textbf{Answer: Girl Scout Cookies}} \\ Clues: A group from Muskogee, Oklahoma is believed to be the first to produce and sell these items popularly sold as a fundraiser for an organization founded by Juliette Gordon Low in 1912. \\

		\underline{\textbf{Answer: {Odysseus}}} \\ Clues: This man's dog Argus dies atop a refuse heap. He reveals himself to a foot-washing maid, Eurycleia (“your-ee-CLAY-uh”). The Laestrygones (“LAY-strih-GOAN-ees”) destroy many ships belonging to his fleet, and he also visits the land of the lotos (“lotus”) -eaters. He kills his wife's suitors with the help of his son, Telemachus (“TELL-uh-MOCK-us”), then reunites with that wife, Penelope. For 10 points, an epic by Homer describes what man's twenty-year quest to get home after the Trojan War?\\

		\underline{\textbf{Answer: {Alice}}} \\ Clues: This character watches a lion and a unicorn fight over a crown, and although her cat Dinah will not talk to her, the Tiger Lily and the other flowers will. She shrinks after drinking a potion labeled "Drink Me," and attends a tea party with a sleepy Dormouse, a March Hare, and a Mad Hatter. \\

		\underline{\textbf{Answer: {Trojan War}}} \\ Clues: Neoptolemus killed King Priam in the final stages of this event, after which Aeneas fled with his son. This event began after the Judgement of Paris and (*) Helen's abduction from King Menelaus of Sparta. After nine years, it finally ended after Greek soldiers got past enemy gates while hiding in a giant wooden horse. For 10 points, name this conflict in Greek mythology that featured warriors like Hector and Achilles.\\

		\underline{\textbf{Answer: {Noah}}} \\ Clues: Seven laws that apply to non-Jews are named for this figure, whose nakedness was uncovered by one of his sons. An agreement this figure made with God is symbolized by the rainbow. He was the son of Lamekh (LAH-meck) and had three sons, Japheth (JAY-feth), Ham, and Shem. To confirm that one of his jobs was complete, he sent a dove to check for dry land. For 10 points, identify this Biblical character who took two animals of each kind in his ark.
        \end{tcolorbox}
        
        \caption{Examples of questions from different clusters.}
        \label{fig:appendix-question-examples-cluster-9}
    \end{figure*}

\begin{figure*}[ht]
        \centering
    
        \begin{tcolorbox}[colback=veasy!5!white,colframe=veasy!75!black,title=Sci. History (V.Easy), fonttitle=\bfseries\small\ttfamily, coltitle=white, fontupper=\tiny\ttfamily]
    		\underline{\textbf{Answer: {Andes} Mountains}} \\ Clues: This mountain range includes the Vilcabamba (“VEEL-cuh-BOM-buh”) sub-range and contains a plateau called the altiplano (“ALL-tee-PLAN-oh”). \\

		\underline{\textbf{Answer: London}} \\ Clues: Hampstead Heath and Kensington Gardens are parks in this city which is served by the "Jubilee Line," "Piccadilly Line," and "Victoria Line" of its subway system, the Underground. A Norman castle built by William the Conqueror is this city's "Tower." \\

		\underline{\textbf{Answer: {Amazon} River}} \\ Clues: The island of Marajo (mah-RAH-hoh) is located at the mouth of this river which was named by Spanish conquistador Francisco de Orellana (day OH-ray-YAH-nah) for the warrior women of Greek mythology. \\

		\underline{\textbf{Answer: {Panama} Canal}} \\ Clues: Lake Gatun (“GAH-tune”) is part of this waterway, whose construction was made possible by the Hay-Bunau-Varilla (“HAY boo-NOW vah-REE-uh”) Treaty and the secession of a province from Colombia. A 1977 agreement between Omar Torrijos (“torr-EE-hos”) and Jimmy Carter resulted in the return of the special zone associated with it. \\

		\underline{\textbf{Answer: Antarctica}} \\ Clues: This geographical feature has its lowest point at Bentley Trench. A lake here lies under Vostok Station. Mt. Erebus is found on Ross Island off itscoast, between Marie Byrd and Victoria lands. The Sentinel Range of the Ellsworth Mountains contains its highest peak, Vinson Massif, located on the Ronne (*) Ice Shelf. \\

		\underline{\textbf{Answer: Saturn}} \\ Clues: Great White Spots are frequent storms on this planet. Its moons include Iapetus, Rhea, Enceladus, and the only known one to have an atmosphere. This planet is less dense than water. The Cassini Division is located in its extensive ring system. For 10 points, name this second largest planet in the solar system, the sixth from the Sun.\\

		\underline{\textbf{Answer: New York City}} \\ Clues: A museum branch located in this city's Fort Tryon Park containing medieval art is known as The Cloisters. One of its straits, which includes Roosevelt Island and Rikers Island, is the East River. \\

		\underline{\textbf{Answer: {Panama} Canal}} \\ Clues: Lake Gatun (“GAH-tune”) is part of this waterway, whose construction was made possible by the Hay-Bunau-Varilla (“HAY boo-NOW vah-REE-uh”) Treaty and the secession of a province from Colombia. \\

		\underline{\textbf{Answer: {Vienna,} Austria}} \\ Clues: This city contains the neo-gothic Votive Church, and its Karlskirche (KARLS-keer-kuh) is the largest Baroque Cathedral north of the Alps. It is the capital of a country with such states as Burgenland, Tyrol, and Styria. This city's Ring Boulevard was ordered to be restructured by Franz Joseph I, and it lies on the Danube just upriver from Bratislava, the capital of Slovakia. \\

		\underline{\textbf{Answer: Orion}} \\ Clues: This constellation contains the Trapezium Cluster and is the site of a late-October meteor shower. One of its stars, formerly known as the Amazon Star, is Bellatrix, and its brightest stars are Betelgeuse and Rigel. Its namesake nebula joins with Hatysa and other stars to form its sword, while Alnitak, Alnilam, and Mintaka form its belt. 
        \end{tcolorbox}
        
        \caption{Examples of questions from different clusters.}
        \label{fig:appendix-question-examples-cluster-10}
    \end{figure*}

\begin{figure*}[ht]
        \centering
    
        \begin{tcolorbox}[colback=veasy!5!white,colframe=veasy!75!black,title=Cult History (V.Easy), fonttitle=\bfseries\small\ttfamily, coltitle=white, fontupper=\tiny\ttfamily]
    		\underline{\textbf{Answer: {Michelangelo} di Lodovico {Buonarroti} Simoni}} \\ Clues: This artist's statues of a dying slave and a horned Moses were to adorn the tomb of Julius II. His only signed work is one in which Mary holds the dead body of Jesus, entitled Pietá (“pee-AY-tuh”). One of his works depicts a nude giant killer holding a sling. \\

		\underline{\textbf{Answer: Charles {Dickens}}} \\ Clues: This author wrote about the eviction of Nell Trent and her grandfather from The Old Curiosity Shop. In another work by this author, Abel Magwitch raises a fortune for the orphan Pip, who loves Estella. He also wrote about Sydney Carton sacrificing himself to save Charles Darnay in a work set in London and Paris. \\

		\underline{\textbf{Answer: Oklahoma}} \\ Clues: This modern state's panhandle was crossed by the Cimarron Cutoff, a branch of the Santa Fe Trail. A city in this state is called "Broken Arrow" because it was settled by Creek people, while part of this state was known as the "Indian Territory." White settlers who anticipated an 1889 decision to open its lands to homesteaders gave this state its nickname: the Sooner State. For 10 points, Tulsa is located in what state between Texas and Kansas?\\

		\underline{\textbf{Answer: {Blessed Virgin Mary}}} \\ Clues: In the Gospel of James, this Biblical figure is described as the child of Anna and Joachim. At the First Council of Ephesus, this figure was given the epithet Theotokos, or "God-Bearer." Martin Luther described this person as "the highest woman." This woman is held to be free from original sin under the doctrine of Immaculate Conception. For 10 points, name this mother of Jesus of Nazareth.\\

		\underline{\textbf{Answer: {Frankenstein,} or the {Modern Prometheus}}} \\ Clues: The protagonist of this work returns home from the University of Ingolstadt to find that Justine Moritz has been accused of his brother William's murder. The title character, whom Robert Walton discovers in the Arctic in a frame story, had earlier married Elizabeth Lavenza, who was killed on their wedding night. \\

		\underline{\textbf{Answer: Paul {Ryan}}} \\ Clues: This politician claimed that he went into politics because of Ayn Rand and made Atlas Shrugged required reading for his staff, but he later said he rejected Rand's atheism. He is the current chair of the House Budget Committee, and one of his budget proposals was titled (*) "The Path to Prosperity." For 10 points what Wisconsin Republican was Mitt Romney's Vice Presidential nominee in the 2012 election?\\

		\underline{\textbf{Answer: {cerebrum}}} \\ Clues: This structure is divided into Brodmann areas, and develops from the telencephalon ("TEAL"-en- SEFF-ah-"lawn"). The corpus callosum ("CORE"-puss kuh-LOE-sum) connects the two hemispheres of this structure, which is divided into temporal, parietal, occipital, and frontal lobes. \\

		\underline{\textbf{Answer: {Michelangelo} di Lodovico {Buonarroti} Simoni}} \\ Clues: This artist's statues of a dying slave and a horned Moses were to adorn the tomb of Julius II. \\

		\underline{\textbf{Answer: {John Quincy Adams}}} \\ Clues: This person negotiated a treaty that ceded Florida to the United States with Luis de Onis (loo-EES day oh-"NIECE") while serving as James Monroe's Secretary of State. This man agreed to name Henry Clay Secretary of State in order to break a deadlock in the House of Representatives; that decision was the first "corrupt bargain." \\

		\underline{\textbf{Answer: Sarah {Palin}}} \\ Clues: This person's visit to Fort Bragg caused a stir when the press was denied entry to a book tour for Going Rogue. This person resigned from the position of Governor of the state closest to Russia shortly after a campaign loss in the most recent general election. Tina Fey did a notable impression of, for 10 points, what unsuccessful vice presidential candidate who ran alongside John McCain in 2008?
        \end{tcolorbox}
        
        \caption{Examples of questions from different clusters.}
        \label{fig:appendix-question-examples-cluster-11}
    \end{figure*}


\end{document}